\documentclass[sigconf, screen]{acmart}

\AtBeginDocument{%
  }

\renewcommand\footnotetextcopyrightpermission[1]{}

\acmConference[MM '26]
  {The 34th ACM International Conference on Multimedia}
  {November 10--14, 2026}
  {Rio de Janeiro, Brazil}

\acmYear{2026}
\copyrightyear{2026}

\usepackage{wrapfig}
\usepackage{enumitem}
\usepackage{bm}
\usepackage{CJKutf8}

\usepackage{lipsum}
\usepackage{multirow}
\usepackage{subcaption} 
\usepackage{graphicx}
\usepackage{arydshln}
\usepackage{paralist}
\usepackage{booktabs}
\usepackage{multirow}
\usepackage{colortbl}
\usepackage[table]{xcolor}
\usepackage{makecell}
\usepackage{array}
\usepackage{placeins}
\usepackage[normalem]{ulem} 

\usepackage{pifont}  

\usepackage{titletoc}

\definecolor{humanblue}{HTML}{1E90FF}   
\definecolor{groupgray}{gray}{0.93}
\definecolor{myblue}{RGB}{227,245,252}
\definecolor{myyellow}{RGB}{252,244,231}
\definecolor{myred}{RGB}{249,236,233}
\definecolor{mygreen}{RGB}{227,253,235}
\definecolor{imppurple}{RGB}{240, 232, 250}
\definecolor{imppos}{RGB}{220, 30, 30}
\definecolor{impneg}{RGB}{50, 180, 50}
\newcommand{\inc}[1]{\textcolor{imppos}{\scriptsize\,+#1}}

\begin{document}

\settopmatter{
  printacmref=false,
  printfolios=true
}

\makeatletter
\renewcommand{\@copyrightpermission}{}
\makeatother

\title{HoloGeo: Mitigating Landmark Bias in Geo-localization via Evidence-Driven Reasoning}

\author{Pengcheng Zhou}
\affiliation{%
  \institution{National University of Singapore}
  \city{Singapore}
  \country{Singapore}}
\email{e1554357@u.nus.edu}

\author{Xuanyu Liu}
\affiliation{%
  \institution{Shandong University of Science and Technology}
  \city{Qingdao}
  \country{China}}
\email{202211081013@sdust.edu.cn}

\author{Yanchen Yin}
\affiliation{%
  \institution{Shandong University of Science and Technology}
  \city{Qingdao}
  \country{China}}
\email{yyc@sdust.edu.cn}

\author{Bobo Li}
\affiliation{%
  \institution{National University of Singapore}
  \city{Singapore}
  \country{Singapore}}
\email{libobo@nus.edu.sg}

\author{Shengqiong Wu}
\authornote{Corresponding author.}
\affiliation{%
  \institution{University of Oxford}
  \city{Oxford}
  \country{United Kingdom}}
\email{shengqiongwu@gmail.com}

\author{Mong-Li Lee}
\affiliation{%
  \institution{National University of Singapore}
  \city{Singapore}
  \country{Singapore}}
\email{dcsleeml@nus.edu.sg}

\author{Wynne Hsu}
\affiliation{%
  \institution{National University of Singapore}
  \city{Singapore}
  \country{Singapore}}
\email{dcshsuw@nus.edu.sg}

\begin{abstract}
Recent advances in Vision-Language Models (VLMs) have significantly improved image geo-localization, yet existing models remain susceptible to \textbf{landmark bias}, causing them to overlook geographical cues or form spurious correlations, ultimately resulting in inaccurate localization. 
To systematically investigate this issue, we first design two quantitative metrics, \textbf{Bias Intensity (BI)} and \textbf{Bias Harmfulness (BH)}, to characterize the impact of landmarks exerted on model reasoning, and establish a comprehensive benchmark, \textbf{LandmarkBias-3K}. 
To mitigate landmark bias, we further propose an evidence-driven reasoning framework, \textbf{HoloGeo}, to improve the reliability of geo-localization. 
HoloGeo is supported by a high-quality dataset, \textbf{BF-30k}, annotated with structured multi-evidence \underline{b}ias-\underline{f}ree reasoning chains. 
By incorporating multi-dimensional rewards, HoloGeo explicitly encourages balanced attention over diverse visual cues and achieves evidence-driven joint reasoning. 
Extensive experiments demonstrate that HoloGeo not only maintains excellent performance on IM2GPS3K and YFCC4k but also significantly outperforms existing open-source VLMs on LandmarkBias-3K, validating its effectiveness for robust geospatial reasoning.
Project page: \url{https://hologeo.github.io/}.
\end{abstract}


\maketitle

\section{Introduction}
\label{sec:intro}

Image geo-localization aims to infer geographic coordinates or specific shooting locations from visual content, and plays a crucial role in a wide range of applications such as autonomous navigation, social media geotagging, and crisis response~\cite{chalvatzaras2022survey, shen2017streetvizor,dutta2024multiview,vivanco2023geoclip}. 
Compared to traditional classification- or retrieval-based methods that rely on fixed grid partitioning or large-scale reference databases~\cite{li2024georeasoner, jia2024g3, muller2018geolocation}, Large Vision-Language Models (VLMs)~\cite{Qwen2.5-VL,chen2024internvl} have shown remarkable progress in geo-localization by leveraging strong cross-modal understanding and integrated world knowledge.
Beyond predicting final locations, emerging systems~\cite{globe2025,wang2025gre,xu2026unlocking,kuckreja2024geochat} further attempt to generate interpretable reasoning trajectories, typically by identifying salient landmarks (e.g., iconic buildings) and then mapping them to geographic hypotheses, moving from black-box prediction toward explainable decision-making.

\begin{figure}[!t]
  \centering
    \includegraphics[width=0.99\linewidth]{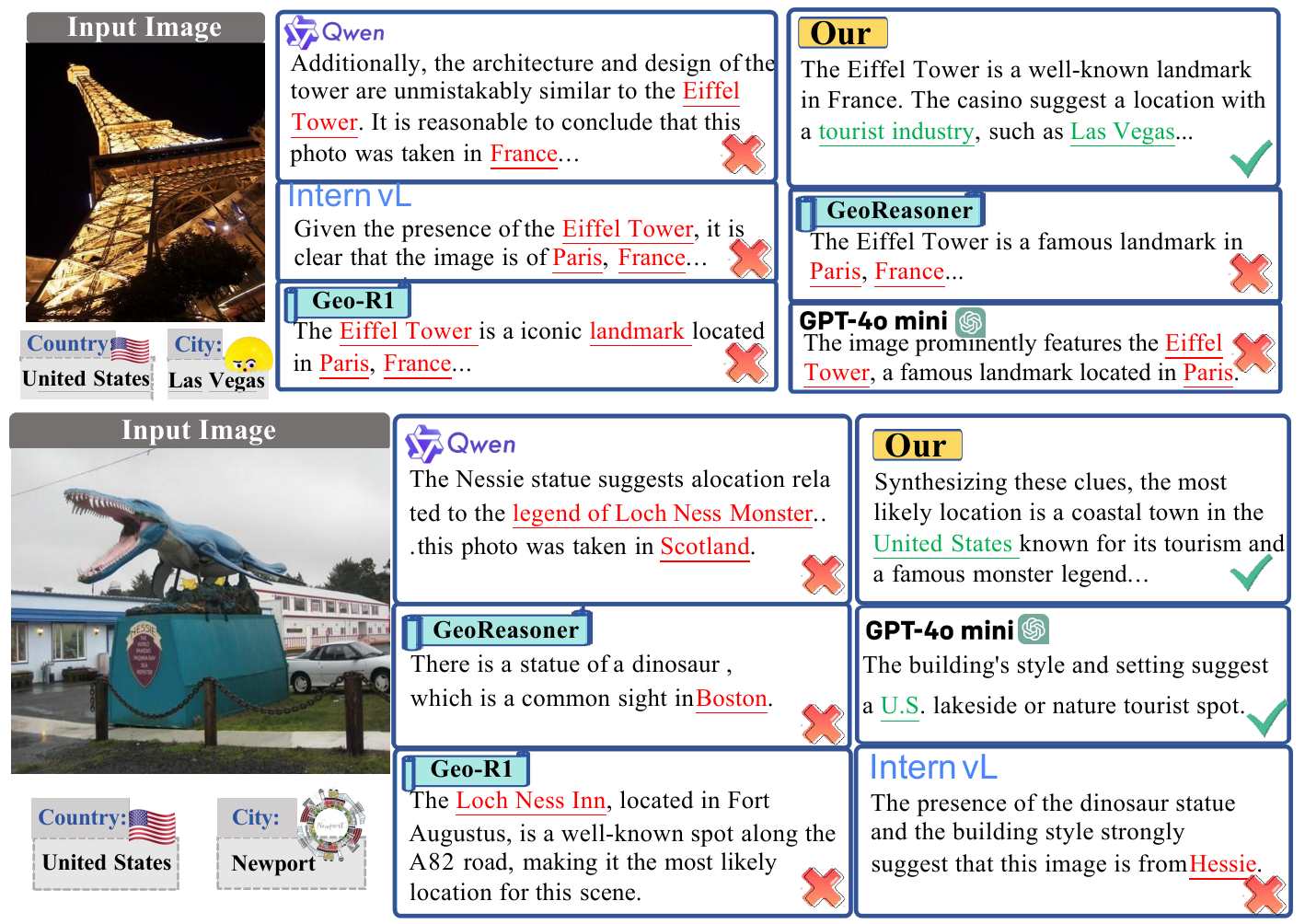}
   \vspace{-2mm}
   \caption{Illustration of landmark bias in geospatial reasoning. Existing models over-rely on salient landmarks (e.g., Eiffel Tower-like structure, Loch Ness Monster statue) and ignore complementary cues, leading to erroneous localization (actual locations: Las Vegas, USA; Newport, USA).} 
   \label{fig:intro}
   \vspace{-2mm}
\end{figure}

Despite progress, existing VLM-based methods still face a critical challenge: \textit{\textbf{landmark bias}}, in which models may over-rely on visually salient landmark-like cues and become over-confident even when such cues are ambiguous, replicated, or weakly informative. 
This bias is amplified by the strong saliency and spurious correlations learned from web-scale data, causing the model to treat a single landmark as sufficient evidence while under-utilizing complementary geographic cues, as the \textit{Eiffel Tower} or the \textit{Loch Ness} in Figure~\ref{fig:intro}. 
Importantly, recent reasoning-oriented approaches~\cite{xu2026unlocking,li2024georeasoner,globe2025} still lack explicit evidence reliability calibration and mechanisms to discourage single-cue dominance; consequently, generated rationales may be fluent yet weakly grounded, failing to guarantee reliable localization. 
Moreover, conventional evaluations largely focus on top-line accuracy, leaving it unclear \emph{how strongly} landmarks dominate the reasoning process and \emph{how harmful} this dominance becomes under landmark ambiguity.
This highlights the need for more fine-grained diagnostic tools to systematically analyze and address landmark bias.

To enable systematic investigation of landmark bias, we first formalize this phenomenon from a reasoning perspective. 
Formally, landmark bias refers to the tendency of models to over-rely on visually salient landmark cues during geospatial reasoning by assigning disproportionate weight to such cues at the expense of underutilizing complementary geographic features, thereby leading to over-confident yet erroneous predictions when landmarks are ambiguous, replicated, or misleading.
Grounded in this formulation, we propose two quantitative metrics, \textbf{Bias Intensity (BI)} and \textbf{Bias Harmfulness (BH)}, to measure the dominance of landmark evidence and its adverse impact under ambiguity. 
BI measures the degree to which landmarks anchor model predictions by evaluating the confidence shift between original and landmark-occluded images;
BH measures the extent to which landmarks mislead the model's preference for the ground truth.
To facilitate fine-grained evaluation, we introduce \textbf{LandmarkBias-3K}, a challenging benchmark specifically curated with scenarios where landmark cues are insufficient or deceptive, thereby exposing the inherent vulnerabilities in biased reasoning behaviors.

To mitigate landmark bias, we further propose \textbf{HoloGeo}, a novel evidence-driven framework that explicitly enforces holistic multi-cue aggregation for geo-localization rather than landmark-centric shortcuts. 
Firstly, we construct a high-quality multi-evidence reasoning dataset, \textbf{BF-30K}, by decomposing images into semantically meaningful regions, organizing geographically grounded analyses into structured multi-evidence reasoning chains, and further refining it through cross-model verification and manual inspection to reduce reasoning errors and hallucinated explanations~\cite{guo2025deepseek, zheng2023judging, huanglarge}. 
Building on this dataset, HoloGeo incorporates multi-dimensional rewards over \textit{prediction correctness} and \textit{evidence coverage}, explicitly encouraging balanced attention across heterogeneous geographic cues. 
Through this design, the model learns to jointly evaluate diverse evidence, align intermediate reasoning with final predictions, and avoid over-reliance on isolated landmark cues.

Extensive experiments on standard benchmarks and custom bias benchmarks verify the effectiveness of the proposed method. Existing general-purpose vision-language models and mainstream geo-localization methods perform poorly on LandmarkBias‑3K, with most achieving below 20\% city-level accuracy; even the state-of-the-art domain-specific model only reaches 23.57\%. In contrast, our proposed HoloGeo attains 27.27\% city-level accuracy on this dataset while maintaining strong performance on conventional geo-localization benchmarks including IM2GPS~\cite{hays2008im2gps}, IM2GPS3K~\cite{vo2017revisiting}, and YFCC4K~\cite{vo2017revisiting}, which fully demonstrates its effectiveness. 
Through saliency map visualization, we find that existing models suffer from landmark bias due to over-reliance on salient landmarks and neglect of broader geographic context, while HoloGeo mitigates this issue via balanced attention distribution. Finally, this validates its capability to effectively mitigate landmark bias while enhancing the robustness of geospatial reasoning.

In all, our contribution can be summarized in threefold:

\begin{compactitem}
\item We present the first systematic study on landmark bias in geospatial reasoning, along with a quantitative metric suite (BI and BH) for fine-grained analysis.

\item We introduce LandmarkBias-3K and BF-30K, the first benchmark and training dataset specifically designed to evaluate and mitigate landmark bias with structured multi-evidence reasoning for geo-localization tasks.

\item We propose HoloGeo, an evidence-driven framework, which encourages balanced multi-cue reasoning through multi-dimensional reward design, effectively reducing landmark bias while preserving strong geo-localization performance.

\end{compactitem}

\section{Related Work}
\label{sec:related_work}

\noindent \textbf{Image Geo-localization.}
Image geo-localization aims to predict the geographic location of a given image and has been widely studied in urban analysis~\cite{kuckreja2024geochat,YanWZCCWZL24,zhang2025geor1}, individual trajectories~\cite{ChengWHMW22}, and geospatial
data mining~\cite{liang2018geoman,PanLW00Z19,WuGCY24,FengZXDM25}. 
Early approaches primarily rely on \emph{retrieval-based} methods~\cite{yang2021cross,zhu2022transgeo,wang2023fine,vivanco2023geoclip,haas2024pigeon,haas2023learning,xia2025fg}, which match query images against large-scale reference databases, and \emph{classification-based} methods~\cite{weyand2016planet,CPlaNet2018,muller2018geolocation,pramanick2022world,clark2023we}, which discretize the earth into predefined regions~\cite{hays2008im2gps, weyand2020GLDv2}. 
While effective, some research~\cite{jia2024g3,zhou2024img2loc,li2024georeasoner} has demonstrated that retrieval-based methods require a large reference database, and classification-based approaches are limited by coarse spatial granularity.
To overcome these limitations, recent works explore \emph{generation-based} approach\-es~\cite{jia2024g3,zhou2024img2loc,dufour2025around} that directly predict geographic coordinates or location descriptions, improving flexibility and scalability. 
With the emergence of large vision-language models (VLMs)~\cite{Qwen2.5-VL,zhu2025internvl3,achiam2023gpt}, geo-localization has further benefited from strong cross-modal understanding and embedded world knowledge, enabling more generalizable and semantically rich predictions~\cite{liu2024image,kuckreja2024geochat}.
More recently, researchers~\cite{cheng2025geoguess,masis2026coordinates,li2024georeasoner} have focused on enhancing the \emph{reasoning capabilities} of VLM-based geo-localization systems. 
Techniques such as chain-of-thought prompting~\cite{li2025pixels,wang2025gre,yerramilli-etal-2025-geochain}, agent-based reasoning~\cite{campos2026gaea,han2025swarm}, and reinforcement learning~\cite{zhang2025geor1,xu2026unlocking,globe2025} have been introduced to generate interpretable reasoning trajectories and improve decision quality. 
However, despite improved reasoning interpretability, existing methods still lack mechanisms to ensure \emph{balanced and reliable evidence utilization}. 
In particular, they tend to over-rely on visually salient landmark cues while under-utilizing complementary geographic signals, leading to biased and over-confident predictions. 
This limitation motivates our work on systematically analyzing and mitigating landmark bias in geo-localization.

\begin{figure*}[!t]
  \centering
   \includegraphics[width=0.98\linewidth]{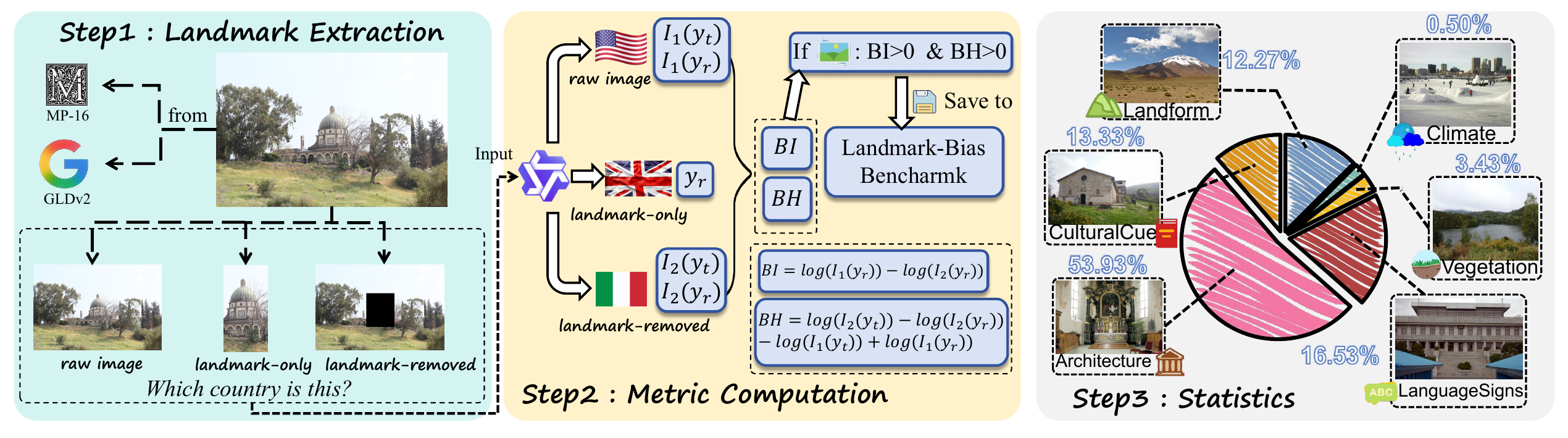}
   \vspace{-2mm}
   \caption{Pipeline for constructing LandmarkBias-3K. 
For each image, landmark regions are first extracted to derive landmark-only and landmark-removed views. 
Model probabilities under the three views are then used to compute Bias Intensity (BI) and Bias Harmfulness (BH), measuring the strength of landmark influence and its harmful effect on the ground-truth prediction, respectively. 
Samples with BI and BH above predefined thresholds are retained to form LandmarkBias-3K. 
The rightmost sub-figure shows the landmark category distribution of the benchmark.}
   \label{fig:landmark_bias_benchmark}
\end{figure*}

\noindent \textbf{Bias and Robustness in VLMs.}
Recent studies have revealed that VLMs are prone to various forms of bias and unreliable reasoning behaviors~\cite{vo2025vision}. 
In particular, models often rely on \emph{spurious correlations} or dataset-specific shortcuts, leading to overconfident predictions that are not well grounded in visual evidence. 
Such issues have been extensively studied in the context of hallucination~\cite{abs-2411-04077,LiuFXXSLK025PhD}, shortcut learning~\cite{chi2025chimera,bleeker2024demonstrating}, and visual bias~\cite{vo2025vision}, where models generate plausible yet incorrect outputs due to over-reliance on partial or misleading cues. 
Several works~\cite{wu2025combating,chen2025perturbollava,li2025instruction} have attempted to improve robustness by encouraging better alignment between visual inputs and generated outputs, for example, through improved training objectives, data curation, or reasoning supervision. 
However, these approaches primarily focus on general-purpose robustness or hallucination mitigation, without explicitly modeling how different types of visual evidence contribute to the final decision.
In the context of geo-localization, this limitation becomes particularly critical.
Existing VLM-based approaches often rely heavily on salient landmark cues during reasoning, yet the dominance of such cues and their potential to mislead predictions under ambiguity have not been systematically studied. 
Moreover, prior work lacks quantitative metrics to measure the extent of such bias, as well as dedicated benchmarks to evaluate its impact on reasoning reliability.
In contrast, we provide a principled formulation and introduce quantitative metrics to measure both the strength and harmfulness of landmark-induced bias, along with a dedicated benchmark and mitigation framework.

\vspace{-2mm}
\section{Landmark-Bias Benchmark}

Our goal is to characterize not only whether landmark cues influence model predictions, but more importantly, whether such influence leads to \emph{harmful} geospatial reasoning errors. 
However, not all salient landmark attraction is detrimental.
Thus, a key challenge lies in distinguishing between benign reliance on informative cues and biased over-reliance that misleads prediction.
To address this, we design two complementary metrics that capture both the strength of landmark influence and its adverse impact on prediction correctness. 
Based on the two metrics, we further construct a challenging benchmark that focuses on cases where landmark cues are highly dominant but unreliable.

\subsection{Quantitation Metrics}
Given an input raw image $x$, we first detect its landmark region $r$ using GroundingDINO~\cite{liu2024grounding}.
We then construct a landmark-removed image $x_r$ by masking out the detected landmark region. 
To quantify the influence of landmark cues on geo-localization, we evaluate the model under three complementary input conditions.

For a location label $y$, we first define
\begin{equation}
I_1(y)=P(y\mid x), \quad
I_2(y)=P(y\mid x_r), \quad
I_3(y)=P(y\mid r),
\end{equation}
where $I_1(y)$ is the model's predicted probability for label $y$ given the full image, $I_2(y)$ is the probability given the landmark-removed image, and $I_3(y)$ is the probability given only the landmark region.
This formulation allows us to isolate the contribution of landmark cues from the remaining contextual evidence.
Then, we define $y_t$ as the ground-truth location label, and $y_r$ as the \emph{landmark-induced label} prediction obtained from the landmark region alone: 
\begin{equation}
y_r=\arg\max_y I_3(y) = \arg\max_y P(y|r),
\end{equation}

Building on these definitions, we quantify landmark-induced bias from two complementary perspectives: \textit{intensity} and \textit{harmlessness}. 
The corresponding metrics are defined as follows:

\noindent \textbf{Bias Intensity (BI).}
Bias intensity measures how strongly the presence of the landmark increases the model's preference for the landmark-induced label $y_r$:
\begin{equation}
BI = \log I_{1}(y_r) - \log I_{2}(y_r).
\end{equation}
A positive $BI$ indicates that the landmark increases the model's confidence in $y_r$, suggesting a strong anchoring effect. 
Values close to zero imply limited influence, while $BI<0$ indicates that the landmark suppresses support for $y_r$.

\noindent \textbf{Bias Harmfulness (BH).}
Bias harmfulness measures whether the landmark shifts the model away from the ground-truth label $y_t$ and toward the landmark-induced label $y_r$:
\begin{equation}
BH = \left(\log I_{2}(y_t)-\log I_{2}(y_r)\right)-\left(\log I_{1}(y_t)-\log I_{1}(y_r)\right).
\end{equation}
where $\log I_1(y_t)-\log I_1(y_r)$ reflects the relative advantage of the true label over the landmark-induced label when the full image is observed, while $\log I_2(y_t)-\log I_2(y_r)$ measures the same advantage after removing the landmark.
A positive $BH$ means that the landmark reduces the model's relative preference for the true label and therefore has a more harmful effect on correct geo-localization.

To ensure a stable probability estimation, Qwen2.5-VL-72B~\cite{Qwen2.5-VL} is adopted for all probability estimates of $I_1(\cdot)$, $I_2(\cdot)$, and $I_3(\cdot)$. 
Additionally, we employ InternVL3-78B~\cite{zhu2025internvl3} for cross-validation to alleviate potential biases inherent in a single large vision-language model. The results are analyzed in the appendix~\ref{app_sec:jcyz}.

\begin{figure*}[t]
  \centering
  \vspace{-2mm}
\includegraphics[width=0.99\linewidth]{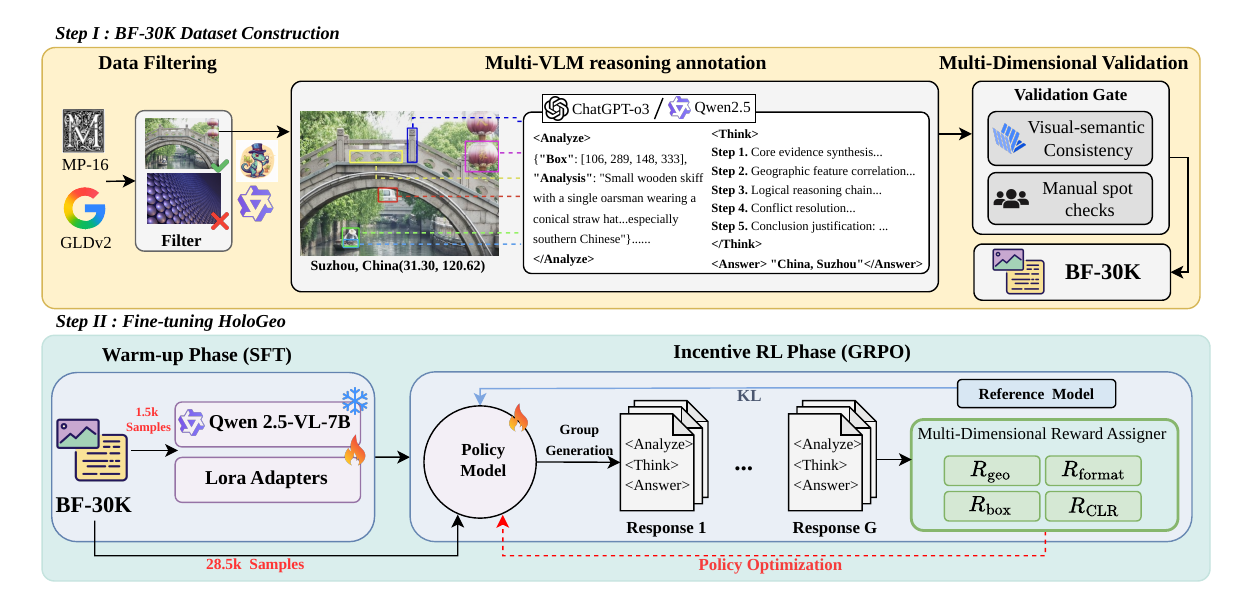}
   \vspace{-5mm}
   \caption{Overview of the proposed HoloGeo. 
   Stage I constructs the BF-30K dataset by filtering samples from {MP-16}~\cite{larson2017benchmarking} and {GLDv2}~\cite{weyand2020GLDv2}, annotating structured multi-evidence reasoning chains, and validating them via cross-model visual-semantic consistency and manual checks.
    In Stage II, built upon Qwen2.5-VL-7B~\cite{Qwen2.5-VL}, HoloGeo adopts a two-stage learning paradigm: a warm-up phase with SFT, followed by GRPO-based reinforcement learning with rewards for geo-location accuracy ($R_{geo}$), visual evidence grounding ($R_{box}$), and comprehensive logical reasoning  $R_{CLR}$).}
   \label{fig:bf_30k}
   \vspace{-2mm}
\end{figure*}

\vspace{-2mm}
\subsection{Benchmark Construction}

Built upon the proposed metrics, we systematically select instances from two large-scale geo-localization datasets, MP-16~\cite{larson2017benchmarking} and GLDv2~\cite{weyand2020GLDv2}, to construct a dedicated benchmark that explicitly targets landmark-induced bias in real-world scenarios. 
As shown in Figure~\ref{fig:landmark_bias_benchmark}, for each image, we first detect its landmark region and construct the landmark-only and landmark-removed views.
Then, we compute the corresponding BI and BH scores, and retain samples with positive BI and BH values that exceed predefined thresholds. 
The detailed threshold settings and selection criteria are provided in Appendix~\ref{app_sec:3k_benchmark}. Finally, this procedure yields \textbf{3,000} images spanning diverse landmark types (e.g., \textit{landform}, \textit{architecture}, \textit{climate}, and \textit{sings}), forming the proposed \emph{LandmarkBias-3K} benchmark.  
Additional dataset statistics are provided in Appendix~\ref{app_sec:3k_benchmark}.

\section{Methodology: HoloGeo}
To mitigate landmark bias and enhance localization robustness, we propose \textbf{HoloGeo}, a unified evidence-driven reasoning framework. 
Technically, we construct the high-quality multi-evidence reasoning-driven dataset, \textbf{BF-30k}, and adopt a two-stage learning paradigm (Figure~\ref{fig:bf_30k}) to guide the model in learning geospatial reasoning patterns, encouraging balanced attention across diverse visual cues and reducing over-reliance on misleading landmarks.

\subsection{BF-30K Dataset Construction}
\label{sec:bf-30k}
Recent works~\cite{xu2026unlocking,zhang2025geor1,globe2025} have explored annotating intermediate reasoning processes to improve model interpretability and performance. 
However, despite considering multiple aspects of an image (e.g., textual cues or scene attributes), the reasoning process may be implicitly anchored to initially salient but potentially misleading cues, introducing early-stage bias and leading to incorrect predictions.
To address this issue, we argue that geospatial reasoning should be grounded in \emph{region-level evidence decomposition} followed by global aggregation~\cite{weyand2016planet,CPlaNet2018}. 
Instead of centering reasoning around a single dominant cue, we propose to construct \emph{structured multi-evidence reasoning chains}, where diverse geographic elements are first analyzed independently and then integrated into a coherent global inference, establishing a complete logical pathway from local visual cues to final localization.
To this end, we design a rigorous construction pipeline, shown in Figure~\ref{fig:bf_30k}, consisting of three stages: data filtering, multi-VLM reasoning annotation, and multi-dimensional validation.

\noindent \textbf{Data Filtering.} 
Large-scale web-crawled datasets provide diverse visual data for geo-localization but often suffer from significant noise~\cite{li2017webvision,wei2022learning,wei2023aggregate}, such as close-up views with insufficient context, generic objects without discriminative localization cues, and images dominated by repetitive or uninformative patterns.
To ensure high-quality training data, 
we design a filtering pipeline to retain samples with reliable geographic signals while preserving diverse landmark bias scenarios. 
Following the previous work \cite{campos2026gaea}, we first adopt GeoCLIP \cite{vivanco2023geoclip} to filter out geographically indistinguishable images from the MP-16~\cite{larson2017benchmarking} dataset, ensuring that all selected samples contain valid geographic cues for geolocation. 
For GLDv2~\cite{weyand2020GLDv2}, we extract the GPS coordinates of each sample via the Wikimedia API~\cite{wikimedia_api}.
We then leverage the geocoder Python library to reversely map these coordinates to the corresponding city and country information. 
We then categorize the filtered samples into three groups: \emph{landmark-biased}, \emph{landmark-containing}, and \emph{non-landmark}. 
For landmark-containing images, we further compute BI and BH scores to identify instances exhibiting strong landmark bias characteristics. 
This results in a balanced dataset covering diverse visual conditions and bias scenarios.

\noindent \textbf{Multi-VLM Reasoning Annotation.}
At this stage, we employ GroundingDINO\cite{liu2024grounding} to detect visual elements in pre-filtered images, automatically identifying bounding boxes for discriminative key regions, thereby providing precise localization for subsequent fine-grained reasoning. We then use Qwen2.5-VL-72B\cite{Qwen2.5-VL} to filter the detected boxes, describing and analyzing each box independently with contextual information to ensure only highly localization-relevant boxes are retained. 
Furthermore, to improve the reliability of geographic element analysis, inspired by recent visual reasoning studies~\cite{man2025argus,zhang2025chain,cheng2025visual}, we feed both the original image and cropped patches of filtered bounding boxes into VLMs for fine-grained annotation, where Qwen2.5-VL-72B and ChatGPT-o3~\cite{openai2024o3o4mini} serve as our annotation models. 
Specifically, we guide the model to complete annotation through a standardized reasoning process: in the \textbf{\texttt{<Analyze>}} stage, the model is required to independently parse the geographic visual content within each valid bounding box for fine-grained feature analysis.
In the \textbf{\texttt{<Think>}} stage, the model aggregates feature information from all bounding boxes, resolves potential ambiguities and contradictions, thereby constructing a unified geographic feature evidence system. Based on this evidence chain, the model performs hierarchical reasoning, gradually narrowing the scope from continent and country to the target city. 
Finally, it outputs the final result in the \textbf{\texttt{<Answer>}} stage, forming a coherent logical connection from local visual cues to global geographic localization.
This design enforces structured multi-evidence reasoning and mitigates over-reliance on single salient cues, providing reliable supervision for geo-localization.

\noindent \textbf{Multi-Dimensional Validation.} 
To ensure data quality, mitigate reasoning trajectory errors, and prevent bias propagation from a single model\cite{guo2025deepseek,wang2025answer}, we design a multi-dimensional validation process. 
We leverage InternVL3-78B \cite{zhu2025internvl3} as an independent reviewer to perform cross-model verification of the annotated data. 
By enforcing visual-semantic consistency across different models, we filter out samples exhibiting hallucinations or logical inconsistencies, thereby improving the reliability of the reasoning chains. 
In addition, we conduct manual spot checks to further identify and remove samples with annotation errors or reasoning contradictions. 
More details are provided in Appendix~\ref{app_sec:bf_30K}. 
The resulting dataset is thus carefully curated to support robust and trustworthy training.

\subsection{Warm-up with SFT}
\label{sec:sft}
To equip the model with preliminary structured geospatial reasoning ability, avoid cold-start issues, and establish a solid foundation for the subsequent reinforcement learning stage~\cite{guo2025deepseek,chu2025sft,zhou2023lima}, we perform SFT using the LoRA strategy~\cite{hu2022lora} for model warm-up. 
Specifically, we randomly sample 1.5K instances from the BF-30k dataset and fine-tune the pre-trained Qwen2.5-VL-7B model.

\subsection{Scaffolding with GRPO}
Drawing inspiration from ~\cite{guo2025deepseek}, we apply GRPO-based RL optimization to further empower geospatial reasoning capability and enhance its generalization~\cite{chu2025sft}.
The pivot part of this stage lies in the design of reward functions, which guide the model toward balanced and reliable reasoning.
Beyond standard format $R_{\text{format}}$ and accuracy-based rewards $R_{\text{geo}}$, we introduce additional objectives to address landmark bias. 
Specifically, to discourage overreliance on salient regions and encourage broader evidence utilization, we design a \emph{visual evidence grounding reward} $R_{\text{box}}$. 
In addition, to ensure the correctness of the reasoning process itself, we introduce a \emph{comprehensive logical reasoning reward} $R_{\text{CLR}}$ that evaluates the consistency and validity of the generated reasoning chains.

\noindent \textbf{Geo-localization Accuracy Reward} \( R_{\text{geo}} \).
This reward evaluates the matching degree between the model's predicted result $\hat{g}_i = (\hat{c}_i, \hat{t}_i)$ and the ground-truth geographic location ${g}_i = (c_i, t_i) $, where $c_i$ and $t_i$ denote the country and city, respectively.
To account for the hierarchical structure of geographic labels, we adopt a hierarchical scoring scheme defined as:
\begin{equation}
R_{\text{geo}}(\hat{g}_i, g_i) = \mathbb{I}[\hat{c}_i = c_i] \cdot \left( \alpha \cdot \mathbb{I}[\hat{t}_i = t_i] + (1-\alpha) \right) 
\end{equation}
where $\mathbb{I}[\cdot]$ is the indicator function and $\alpha \in [0,1]$ controls the weight of city-level accuracy.
The reward is 0 if the country prediction is incorrect. 
If the country is correct but the city is not, a partial reward of $1-\alpha$ is assigned. 
The full reward of 1.0 is obtained only when both country and city predictions are correct. 
This design encourages the model to first capture coarse-grained geographic information before refining fine-grained localization. 
In our experiments, we set $\alpha = 0.8$.

\noindent \textbf{Format Reward} \( R_{\text{format}} \).
This reward ensures the model's output contains the \texttt{<Analyze>}, \texttt{<Think>}, and \texttt{<Answer>} structure: a score of 1.0 is awarded for compliance, and 0 otherwise.

\noindent \textbf{Visual Evidence Grounding Reward} \( R_{\text{box}}\).
This reward measures how well the model grounds its reasoning in relevant visual evidence by evaluating the alignment between predicted and ground-truth bounding boxes, so as to encourage the model to cover diverse geographic cues rather than over-relying on a single salient region.
For the $i$-th sample, let \( \hat{B}_i \) denote the set of predicted bounding boxes parsed by the model from its generation module, and \( B_i \) denote the set of ground-truth annotated boxes for key regions corresponding to the image.
We adopt an IoU-based matching criterion with threshold $\tau_{\text{IoU}}$. 
Specifically, for each ground-truth box $b_k \in B_i$, we compute its maximum overlap with predicted boxes.
A ground-truth box is considered successfully covered if the maximum IoU exceeds $\tau_{\text{IoU}}$. 
The reward is defined as:
\begin{equation}
R_{\text{box}}=\frac{1}{|B_i|}\sum_{b_k \in B_i} \mathbb{I}\left(\max_{b_{\hat{j}} \in \hat{B}_i} \text{IoU}(b_{\hat{j}}, b_k) \geq \tau_{\text{IoU}}\right),
\end{equation}
where \( \mathbb{I}(\cdot) \) is the indicator function.
This reward score is positively correlated with the ratio of successfully matched ground-truth boxes, encouraging the model to accurately locate and cover key geographic regions in images.

\begin{table*}[!t]
\centering
\caption{\textbf{Comparison with existing geo-localization models.} Evaluation on IM2GPS, IM2GPS3k, and YFCC4k benchmarks. Values indicate accuracy (\%) at different distance thresholds.}
\label{tab:main_results}
\fontsize{8}{9}\selectfont
\setlength{\tabcolsep}{3.2mm} 
\begin{tabular}{lccccccccc}
\toprule
\multirow{2}{*}{\textbf{Model}} & \multicolumn{3}{c}{\textbf{IM2GPS}}  & \multicolumn{3}{c}{\textbf{IM2GPS3k}}  & \multicolumn{3}{c}{\textbf{YFCC4k}} \\
\cmidrule(r){2-4} \cmidrule(r){5-7} \cmidrule(r){8-10}
& \textbf{City} & \textbf{Region} & \textbf{Country } & \textbf{City} & \textbf{Region} & \textbf{Country } & \textbf{City} & \textbf{Region} & \textbf{Country } \\
\midrule
\rowcolor{groupgray}\multicolumn{10}{c}{$\bullet$ \textit{Traditional Domain-Specific  Model}} \\
PlaNet~\cite{weyand2016planet} & 24.5 & 37.6 & 53.6 & 24.8 & 34.3 & 48.4 & - & - & - \\
CPlaNet~\cite{CPlaNet2018} & 37.1 & 46.4 & 62.0 & 26.5 & 34.6 & 48.6 & - & - & - \\
ISNs\cite{muller2018geolocation} & 43.0 & 51.9 & 66.7 & 28.0 & 36.6 & 49.7 & - & - & - \\
TransLocator\cite{pramanick2022world} & 48.1 & 64.6 & 75.6  & 31.1 & 46.7 & 58.9 & 18.6 & 27.0 & 41.1 \\
GeoCLIP~\cite{vivanco2023geoclip} & 41.8 & 60.8 & 77.2  & 34.5 & 50.7 & 69.7 & - & - & - \\
GeoDecoder~\cite{clark2023we} & 50.2 & 69.0 & 80.0  & 33.5 & 45.9 & 61.0 & - & - & - \\
PIGEON~\cite{haas2024pigeon} & 40.9 & 63.3 & 82.3  & 36.7 & 53.8 & 72.4 & - & - & - \\
\cdashline{1-10} 
\rowcolor{groupgray}\multicolumn{10}{c}{$\bullet$ \textit{General VLM Model}} \\
InternVL2-8B~\cite{chen2024internvl} & 0.8 & 3.0 & 3.8 & 4.2 & 6.8 & 9.4 & 0.4 & 0.9 & 2.0 \\
LLaVA-Next-Mistral-7B~\cite{liu2024llavanext} & 7.6 & 8.4 & 11.4 & 4.1 & 5.1 & 7.4 & 2.0 & 2.7 & 4.7 \\
Qwen2.5-VL-7B~\cite{Qwen2.5-VL} & 35.0 & 46.8 & 59.1 & 27.9 & 38.9 & 49.9 & 11.5 & 16.6 & 24.1 \\
\cdashline{1-10} 
\rowcolor{groupgray}\multicolumn{10}{c}{$\bullet$ \textit{VLM-based Domain-Specific  Model}} \\
GeoReasoner~\cite{li2024georeasoner} & 24.9 & 48.1 & 65.8 & 26.5 & 40.4 & 57.7 & 6.3 & 10.4 & 17.3 \\
GaGA~\cite{dou2024gaga} & 38.8 & 54.8 & 75.1 & 33.0 & 48.0 & 67.1 & - & - & - \\
GAEA~\cite{campos2026gaea} & 43.0 & 57.4 & \textbf{77.2} & 36.9 & 56.0 & \textbf{73.2} & - & - & - \\
Geo-R1~\cite{xu2026unlocking} & 42.2 & 59.5 & \textbf{77.2} & 35.4 & 52.3 & 69.7 & 17.0 & 29.0 & 48.9 \\
GeoChat~\cite{kuckreja2024geochat} & 38.8 & 59.1 & 76.8 & 34.7 & 51.7 & 69.4 & 17.0 & 29.2 & 49.8 \\
GLOBE~\cite{globe2025} & 44.3 & 59.4 & 76.3 & \textbf{40.2} & \textbf{56.2} & 71.5  & 18.0 & 30.7  & 50.6  \\
\cdashline{1-10} 
HoloGeo-SFT(Ours) & 40.9 & 56.5 & 73.4 & 36.9  & 52.3 & 70.8 & 18.1 & 27.0 & 44.5 \\

HoloGeo(Ours) & \textbf{47.3} & 60.3 & 76.8 &  38.5 & 53.7 & 70.8 & \textbf{18.9 }& \textbf{31.7} & \textbf{51.5} \\
\bottomrule
\end{tabular}

\end{table*}

\noindent \textbf{Comprehensive Logical Reasoning Reward} \( R_{\text{CLR}} \).
This reward evaluates the validity and reliability of the model’s reasoning process, encouraging consistent, well-grounded, and multi-evidence reasoning.
We employ Qwen3.5-35B-A3B~\cite{qwen3.5} as a reward model to score each sample pair $(I_i, \hat{r}_i)$, where $I_i$ is the input image and $\hat{r}_i$ is the generated reasoning trajectory.
Specifically, the reward model evaluates the reasoning according to the structured three-stage process.
\textbf{1) \texttt{<Analyze>} Visual Factuality:} The described visual attributes must be consistent with the content of the corresponding bounding box regions.
\textbf{2) \texttt{<Think>} Logical Consistency:} The reasoning process should be coherent and avoid over-reliance on a single geographic cue, unless no other informative evidence is available.
\textbf{3) \texttt{<Answer>} Answer Specificity:} The predicted location should be expressed as specific country and city names, without vague or ambiguous descriptions.
The reward model outputs the probabilities of \texttt{yes} (valid reasoning) and \texttt{no} (invalid reasoning). 
We define the reward as the normalized probability of valid reasoning:
\begin{equation}
R_{\text{CLR}} = \frac{\mathbb{P}(\text{yes} \mid I_i, \hat{r}_i; \theta_{\text{reason}})}{\mathbb{P}(\text{yes} \mid I_i, \hat{r}_i; \theta_{\text{reason}}) + \mathbb{P}(\text{no} \mid I_i, \hat{r}_i; \theta_{\text{reason}})},
\end{equation}
where \(\theta_{\text{reason}}\) denotes the predictive distribution of the reward model, the numerator is the probability that the reasoning is valid, and the denominator is the sum of the probabilities of validity and invalidity. 
This score serves not only as a soft metric for reasoning validity, but also guides the model to realize visual fact-based multi-element geographic reasoning. 

Finally, the total reward is:
\begin{equation}
r_i =R_{\text{format}} + \lambda_1 R_{\text{geo}} + \lambda_2 R_{\text{box}} + \lambda_3 R_{\text{CLR}},
\end{equation}
where we set the reward weights to \( \lambda_1 = 1.2 \), \( \lambda_2 = 0.3 \), and \( \lambda_3 = 0.5 \).

\noindent\textbf{GRPO-based Reinforcement Learning}.
Based on the reward signals defined above, we adopt the GRPO~\cite{guo2025deepseek} to fine-tune the model $\pi_\theta$ optimized via supervised fine-tuning in Section~\ref{sec:sft} on the BF-30K ($D_{rl}$) dataset. 
GRPO introduces an intra-group normalization mechanism that optimizes the relative preferences $A_i$ among candidate outputs $O = \{o_i\}$ for each input sample $I$, thereby enhancing robustness against fluctuations in reward scales.
A KL-divergence regularization term $\mathbb{D}_{\mathrm{KL}}$ is introduced to prevent the optimized policy $\pi_\theta$ from diverging excessively from the reference model $\pi_{\mathrm{ref}}$:
\begin{equation}
\begin{split}
     \max_{\pi_\theta} &\;
    \mathbb{E}_{[I \sim D_{\text{rl}}, \{o_i\}_{i=1}^{G} \sim \pi_\theta(O|I)}  
    \Bigg[ \frac{1}{G}
        \sum_{i=1}^G 
        \text{min}(\frac{\pi_\theta(o_i)}{\pi_{\theta_{\text{old}}}(o_i)} A_i, \\ &\text{clip}(\frac{\pi_\theta(o_i)}{\pi_{\theta_{\text{old}}}(o_i)},    1-\epsilon, 1+\epsilon) A_i )
        - \beta \, \mathbb{D}_{\mathrm{KL}}\big(\pi_\theta \| \pi_{\mathrm{ref}}\big)
    \Bigg],
\end{split}
\end{equation}
where $\epsilon, \beta$ are the hyper-parameters.

\section{Experiments}
\label{sec:experiments}

\subsection{Setups}
\label{sec:setups}

\noindent \textbf{Datasets.}
We evaluate all models on four geo-localization benchmarks:
\textbf{IM2GPS}~\cite{hays2008im2gps},
\textbf{IM2GPS3K},~\cite{vo2017revisiting}
\textbf{YFCC4K}~\cite{vo2017revisiting},
and our proposed \textbf{Landmark-Bias-3K Benchmark}.
IM2GPS and IM2GPS3K are widely used benchmarks that primarily consist of landmark-centric images, while YFCC4K contains more diverse scenes. In contrast, the Landmark-Bias-3K benchmark is explicitly curated to highlight landmark-induced bias.

\noindent \textbf{Benchmark Models.}
We evaluate a diverse set of geo-localization models, including
(i) classical image-based geo-localization methods,
(ii) recent vision language models,
and (iii) reasoning-oriented geo-localization models.
For classical methods, we include PlaNet\cite{weyand2016planet}, CPlaNet\cite{CPlaNet2018}, ISNs\cite{muller2018geolocation}, TransLocator\cite{pramanick2022world} and so on.
For vision-language models, we evaluate InternVL2-8B\cite{chen2024internvl}, LLaVA-Next-Mistral-7B\cite{liu2024llavanext} and Qwen2.5-VL-7B\cite{Qwen2.5-VL} using their official checkpoints and inference protocols.
We further include several reasoning-oriented models, including GeoReasoner~\cite{li2024georeasoner}, GeoChat~\cite{kuckreja2024geochat}, and Geo-R1~\cite{xu2026unlocking}, among others, which explicitly incorporate structured or multi-step geographic reasoning.

\noindent \textbf{Evaluation Metrics.}
For geo-localization benchmarks, we follow previous work~\cite{li2024georeasoner,xu2026unlocking,jia2025georanker} and report the percentage of predictions whose geographic distance to the ground-truth coordinate falls within fixed thresholds of \textbf{25 km}, \textbf{200 km}, and \textbf{750 km}.
Some models output discrete place names; we convert these predictions into their corresponding geographic center coordinates using external geocoding tools for distance-based evaluation.
All metrics are reported as accuracy percentages, and higher values indicate better geo-localization performance.

\noindent \textbf{Implementations.}
We adopt Qwen2.5-VL-7B-Instruct as the base model and perform parameter-efficient fine-tuning using LoRA~\cite{hu2022lora} with a rank of 64 and an alpha of 128. 
AdamW optimizer is applied with a learning rate of 1e-5 for LoRA parameters, a cosine learning rate scheduler with a warmup ratio of 0.01, a weight decay of 0.1, and a global batch size of 32. 
Details are provided in Appendix~\ref{app_sec:settings}.

\begin{table}[!t]
\centering
\caption{Evaluation results on LandmarkBias-3K. \textcolor{imppos}{+} indicates gains over backbone Qwen2.5-VL-7B~\cite{Qwen2.5-VL}.}
\label{tab:results_on_landmark_bias}
\vspace{-2mm}
\fontsize{8}{9}\selectfont
\setlength{\tabcolsep}{3mm} 
\begin{tabular}{l ccc}
\toprule
\multirow{2}{*}{\textbf{Model}} & \textbf{City} & \textbf{Region} & \textbf{Country } \\
&  (25 km) &  (200 km) & (750 km) \\
\midrule
\rowcolor{groupgray}\multicolumn{4}{c}{$\bullet$ \textit{General VLM Model}} \\
GPT-4o-mini~\cite{openai2024gpt4omini} & 19.47  & 35.77  & 59.00 \\
GPT-4o~\cite{openai2024gpt4o} &  30.53 & 47.50 & 67.83\\
\cdashline{1-4}
 Qwen2.5-VL-7B~\cite{Qwen2.5-VL} & 16.83 & 28.67 & 44.57 \\
InternVL2-8B~\cite{chen2024internvl} & 5.77 & 11.54 & 32.40 \\
LLaVA-Next-Mistral-7B~\cite{liu2024llavanext}  & 0.74 & 1.59 & 5.10 \\
\hline
\rowcolor{groupgray}\multicolumn{4}{c}{$\bullet$ \textit{Domain-Specific Model}} \\
TransLocator~\cite{pramanick2022world} & 18.60 & 27.00 & 41.10 \\
GeoCLIP~\cite{vivanco2023geoclip}  & 17.96 & 30.45 & 60.88\\
GeoDecoder~\cite{clark2023we} & 18.23 & 34.27 & 59.60  \\
GeoChat~\cite{kuckreja2024geochat} & 19.67 & 38.77 & 63.37  \\
GeoReasoner~\cite{li2024georeasoner} & 10.67 & 22.77 & 47.13 \\
GLOBE~\cite{globe2025} & 23.27 & 44.27 & 63.90 \\
Geoagent~\cite{jin2026geoagent} & 23.57 & 45.27 & 67.07 \\
\cdashline{1-4}
HoloGeo-SFT & 20.03 & 30.33 &  64.05 \\
HoloGeo-SFT+RL & \textbf{27.27} & \textbf{47.07} & \textbf{68.20}\\
\specialrule{0em}{-2pt}{-1pt} & (\inc{10.44}) &  (\inc{18.40})  &  (\inc{23.63})  \\
\bottomrule
\end{tabular}

\vspace{-2mm}
\end{table}

\subsection{Main Results}

\noindent \textbf{Results on Existing Benchmarks.}
Table \ref{tab:main_results} summarizes the comparison with existing geo-localization methods. Our SFT-only model already exhibits strong and competitive performance across multiple geographic scales, achieving 40.9\% (City) on IM2GPS and 18.1\% (City) on YFCC4k with more complex and diverse scenes, surpassing the recent GLOBE model~\cite{globe2025}. 
Although the SFT variant does not always yield the highest accuracy among all models, its performance remains consistently close to the state-of-the-art level, indicating that the basic representation and training design are effective.
Notably, further performance improvements are achieved by introducing GRPO-based reinforcement learning. 
Specifically, HoloGeo attains the best accuracy of 47.3\% (City) and 60.3\% (Region) on the IM2GPS dataset and achieves state-of-the-art results across city, region, and country levels on YFCC4k. 
Although its performance on IM2GPS3k is slightly inferior to the state-of-the-art, this is mainly attributed to the severe label distribution redundancy in this test set and the targeted optimization adopted by competing methods. 
In contrast, HoloGeo focuses more on building global generalization ability, thus exhibiting stronger robustness in diverse real-world scenarios.
These improvements stem from HoloGeo’s multi-dimensional reward mechanism and evidence-driven reasoning, which explicitly guide the model to focus on and leverage diverse effective geographic visual elements in images.

\begin{table}[!t]
\centering
\caption{Ablation study on HoloGeo's training components and reward configurations. Evaluated on the IM2GPS dataset with Qwen2.5-VL-7B-Instruct~\cite{Qwen2.5-VL} as the base model.}
\label{tab:ablation_training}
\vspace{-2mm}
\resizebox{\linewidth}{!}{
\begin{tabular}{l c ccc cc}
\toprule
\multirow{2}{*}{\textbf{Models}} & \multirow{2}{*}{\textbf{SFT}} & \multicolumn{3}{c}{\textbf{Reward Function}} & \multirow{2}{*}{\makecell{\textbf{City}}} & \multirow{2}{*}{\makecell{\textbf{Region}}}  \\
\cmidrule(r){3-5}
& & $R_{geo}$ & $R_{box}$ & $R_{CLR}$ &  &   \\
\midrule
Qwen2.5-VL-7B & - & - & - & - & 35.02  & 46.83 \\
Qwen2.5-VL-7B  & \checkmark & - & - & - & 40.93 & 56.54 \\
\midrule
\textit{HoloGeo (Full)} & \checkmark & \checkmark & \checkmark & \checkmark & 47.26 & 60.34 \\
\cdashline{1-7}
\quad w/o $R_{geo}$ & \checkmark & - & \checkmark & \checkmark & 41.35  & 59.49  \\
\quad w/o $R_{box}$ & \checkmark & \checkmark & - & \checkmark &  43.03 & 59.07  \\
\quad w/o $R_{CLR}$ & \checkmark & \checkmark & \checkmark & - & 45.99 & 59.49 \\
\bottomrule
\end{tabular}
}
\vspace{-3mm}
\end{table}

\begin{figure*}[t]
  \centering
  \vspace{-8pt}
  \includegraphics[width=0.95\textwidth]{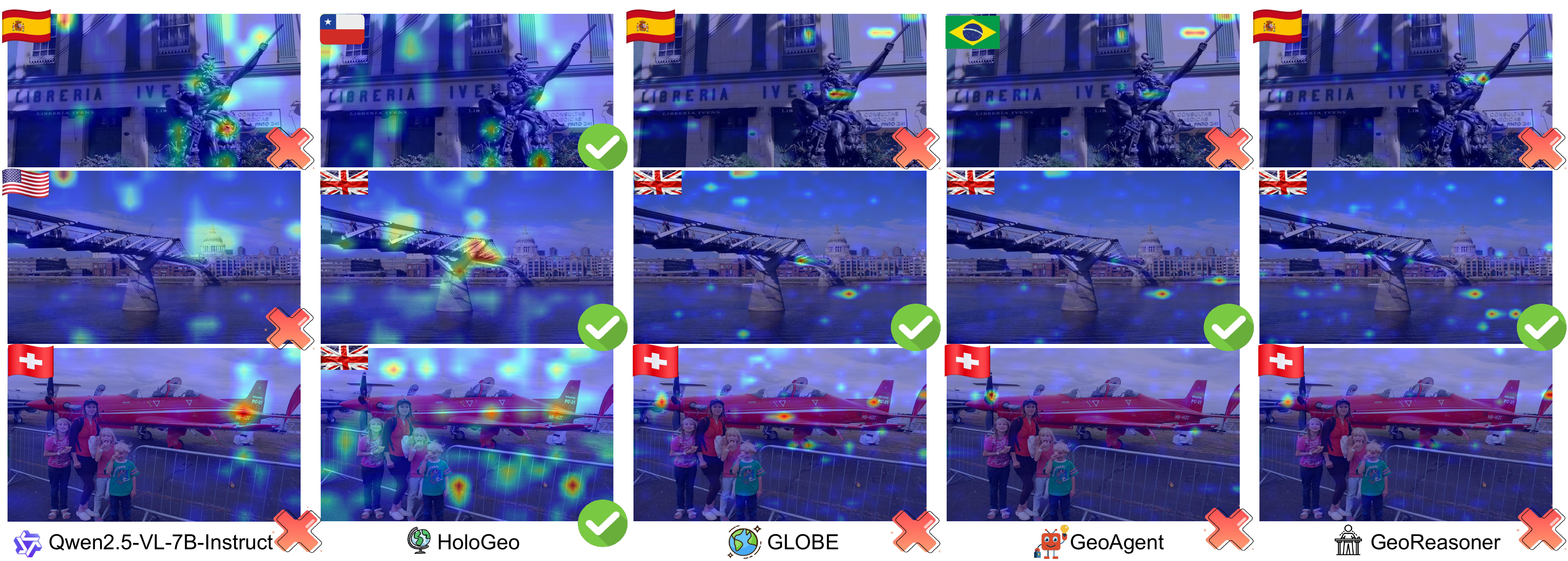}
  \vspace{-13pt}
  \caption{Comparison of Attention Distributions Across Qwen2.5-VL-7B-Instruct, GLOBE, GeoAgent, GeoReasoner, and HoloGeo.}
  \label{fig:reason_landmark}
\end{figure*}

\noindent \textbf{Results on Landmark-bias Datasets.}
Table \ref{tab:results_on_landmark_bias} shows the comparison results of various models on our constructed LandmarkBias-3K benchmark, which is specifically designed to test the reasoning robustness of models when encountering misleading salient landmarks. 
It can be seen that both general VLM models and existing geo-localization methods are affected by landmark bias, resulting in low overall accuracy. 
Among them, models such as InternVL2-8B achieve less than 10\% (City) accuracy. Even the strongest domain-specific models such as GLOBE and Geoagent only reach 23.27\% and 23.57\% (City) accuracy, respectively. 
In contrast, our HoloGeo-SFT model achieves 20.03\% at the city level, demonstrating preliminary anti-bias ability. After incorporating GRPO-based reinforcement learning, HoloGeo-SFT+RL further improves the accuracy to 27.27\% (City), while maintaining leading performance on region-level and country-level metrics. 
These results fully prove that HoloGeo can alleviate over-reliance on a single landmark and conduct robust reasoning by integrating effective geographic visual elements in challenging scenarios with misleading landmarks.

\begin{figure}[!t]
\centering
\includegraphics[width=0.99\linewidth]{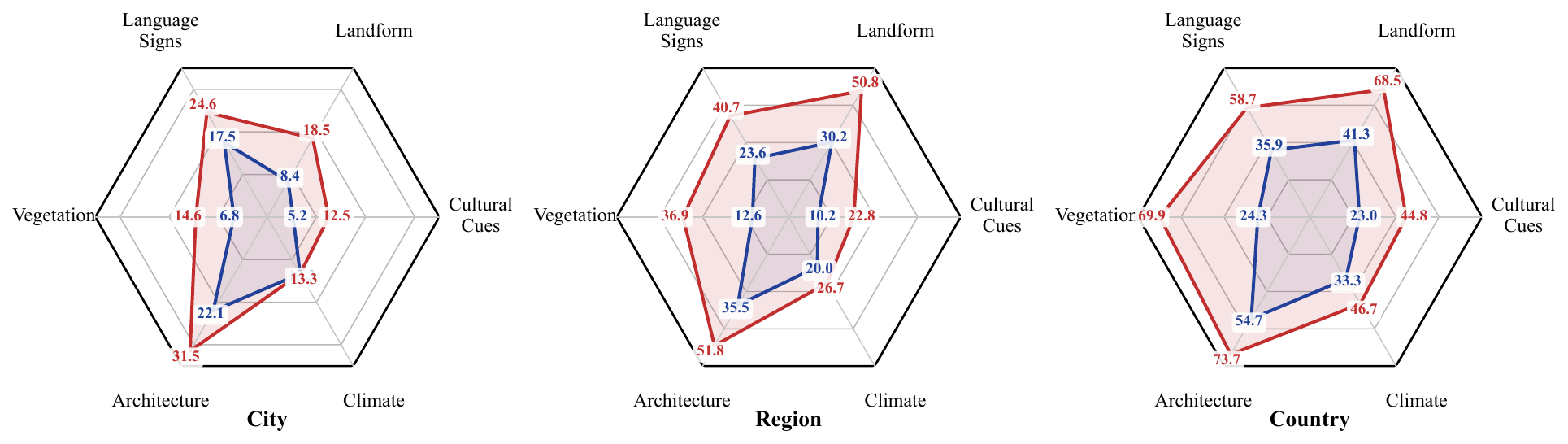}
\vspace{-3mm}
\caption{Prediction of \textcolor[HTML]{204099}{Qwen2.5-VL-7B} and \textcolor[HTML]{c22e2e}{HoloGeo} on different landmark types.}
\label{fig:landmark_type_accuracy}
\vspace{-8pt}
\end{figure}

\subsection{Ablation Study}
In Table~\ref{tab:ablation_training}, we further investigate the impact of different training paradigms and reward configurations on the performance of HoloGeo. Specifically, we compare LoRA SFT with our proposed GRPO-based reinforcement learning approach. For SFT, the city-level accuracy is improved from 35.02\% to 40.93\%, indicating that structured geographic reasoning supervision can endow the model with basic geo-localization capabilities. For GRPO, we conduct a reward ablation study by evaluating different combinations of the three reward components. It shows that GRPO with the full set of rewards achieves the best overall performance of 47.26\% in city-level accuracy. Removing any one of $R_{geo}$, $R_{box}$, and $R_{CLR}$ leads to significant performance drops, which underscores the importance of supervision on both location-level correctness and comprehensive reasoning over multiple visual elements.

\begin{figure}[!t]
  \centering
   \includegraphics[width=1.0\linewidth]{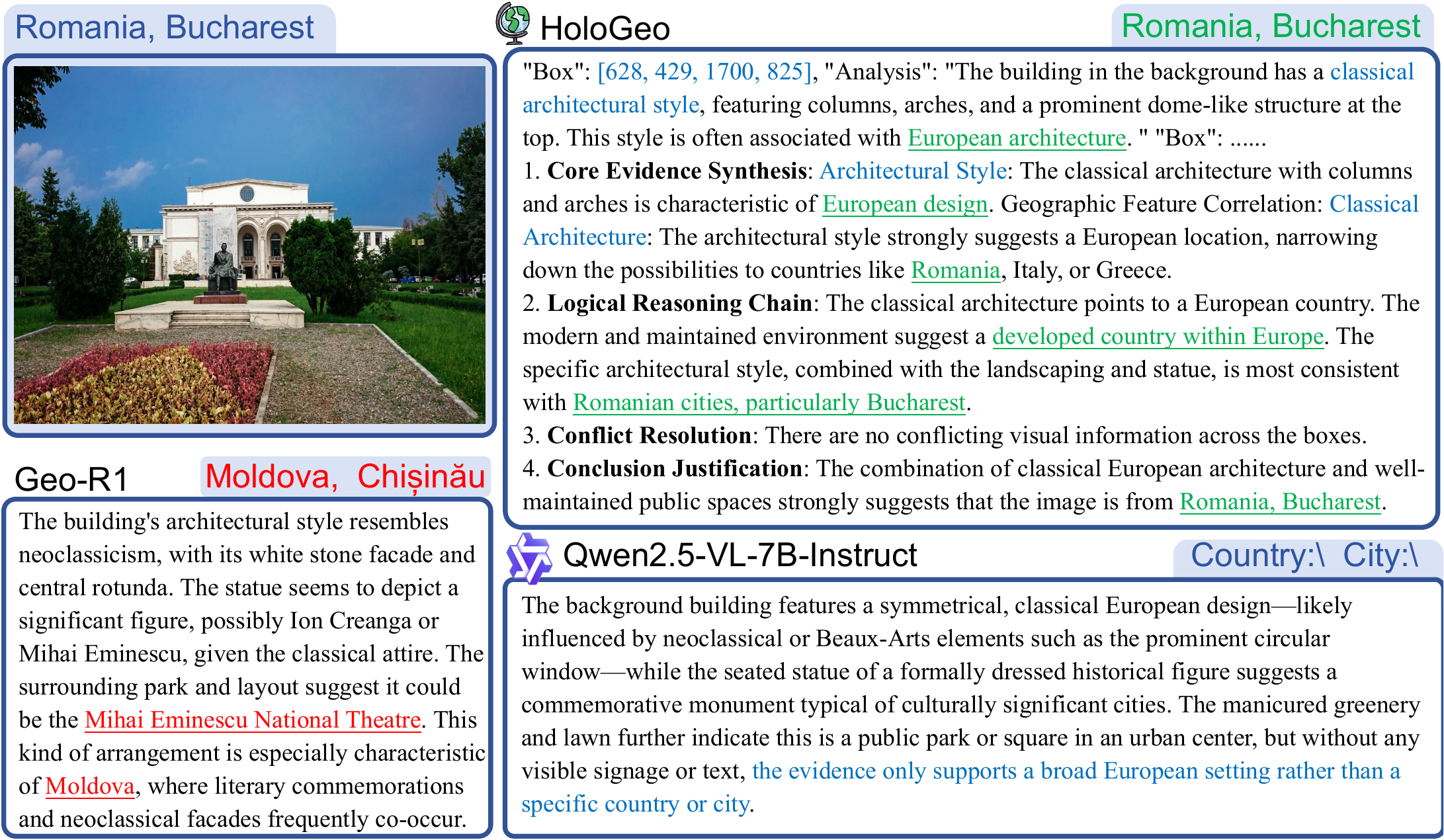}
   \vspace{-6mm}
   \caption{Case study: Reasoning comparison of three different models on the same input image. Reliable geographic information identified by the models is annotated in the text.}
   \vspace{-5mm}
   \label{fig:case_study}
\end{figure}

\subsection{In-depth Analyses}

\noindent \textbf{Why landmark-bias occurs?}
Landmark-induced bias mainly arises from models concentrating visual attention on salient landmarks while ignoring broader geographic context. To this end, we employ gradient-based saliency analysis to generate saliency maps of the original images, which intuitively visualize the key visual regions that models focus on during geographic reasoning. As shown in Figure~\ref{fig:reason_landmark}, for Qwen2.5-VL-7B-Instruct and other VLM-based geolocation models, the saliency maps are highly concentrated on a few visually prominent landmarks (e.g., iconic statues, distinctive buildings, or aircraft), while most surrounding regions of the scene receive little to no attention. Such a landmark-centered attention pattern causes models to overemphasize globally recognizable objects and neglect complementary geographic cues, including environmental context. Consequently, model predictions become strongly anchored to the landmarks themselves, leading to a significantly higher risk of confusion when similar landmarks appear in different locations.  In contrast, HoloGeo fully considers both landmark and non-landmark regions such as surrounding architecture and environmental details, exhibiting a more balanced attention distribution. This holistic attention pattern enables HoloGeo to integrate diverse geographic evidence and reduce over-reliance on any single visual cue. As a result, landmark bias is effectively mitigated, leading to more robust and reliable geolocation performance.

\noindent \textbf{How does landmark bias vary across landmark types?}
Figure~\ref{fig:landmark_type_accuracy} analyzes how landmark-induced bias varies across landmark types by reporting geo-localization accuracy at the city, region, and country levels. The key observation is that the severity of bias depends on how specific and misleading a cue is for localization. 
For Qwen2.5-VL-7B, Architecture and Language Signs show relatively low city-level accuracy but noticeably higher country-level accuracy, suggesting that the model often locks onto globally recognizable styles or text patterns that provide a coarse geographic prior but are not distinctive enough to pin down the exact city, leading to plausible yet incorrect fine-grained predictions. 
In contrast, Climate yields comparatively higher accuracy, which we attribute to climate cues being less shortcut-like: climate-related evidence typically constrains the scene to broad latitudinal or regional bands rather than pointing to a single iconic place, so it induces a milder form of bias and is less likely to anchor the model to a specific wrong location. 
Finally, weak or transferable cues such as Vegetation, Landform, and Cultural Cues result in low accuracy across granularities because these signals are often shared across many regions and require careful integration with surrounding context; when the model’s attention is dominated by any salient element, it struggles to accumulate sufficient discriminative evidence for reliable localization.

\noindent \textbf{Case Study.}
Figure~\ref{fig:case_study} presents a representative case of landmark-induced bias, where salient landmarks lead baseline models to rely on superficial similarity and produce incorrect predictions. In contrast, HoloGeo mitigates this effect by integrating complementary contextual cues beyond the landmark, enabling correct geo-localization. We further provide additional case studies in Appendix~\ref{app_sec:more_case_examples} to qualitatively illustrate how biased landmarks influence model behavior and decision-making.

\section{Conclusion}
This paper addresses the landmark bias issue in image geolocalization by proposing a holistic reasoning-driven framework, HoloGeo.
On the one hand, we introduce two quantitative metrics, Bias Intensity (BI) and Bias Harmfulness (BH), to systematically characterize the impact of landmark cues on model decisions, and construct both a diagnostic evaluation benchmark (LandmarkBias-3K) and a high-quality training dataset (BF-30K) with structured multi-evidence reasoning chains.
On the other hand, to mitigate landmark bias, HoloGeo integrates supervised fine-tuning with GRPO-based reinforcement learning, guided by a multi-dimensional reward design that promotes visual evidence grounding and logical reasoning consistency. 
This framework encourages models to shift from over-reliance on single salient landmarks to balanced reasoning over diverse geographic cues. 
Extensive experiments on {IM2GPS}, {IM2GPS3K}, {YFCC4K}, and LandmarkBias-3K demonstrate that HoloGeo improves both robustness under landmark ambiguity and overall geo-localization performance.
In future work, we plan to explore tool-augmented and agent-based reasoning to incorporate external geographic knowledge, as well as more data-efficient learning paradigms such as few-shot and zero-shot generalization.

\bibliographystyle{ACM-Reference-Format}
\bibliography{acmart}

\appendix

\startcontents[appendix]

\section*{Appendix}

\begingroup
\renewcommand{\contentsname}{}
\printcontents[appendix]{}{1}{}
\endgroup

\section{Detailed Landmark-Bias Quantitation Metrics}
\label{app_sec:metrics}

\subsection{Distribution of BI and BH}
\begin{figure*}[!t]
  \centering
  \vspace{-10pt}
  \includegraphics[width=0.8\textwidth]{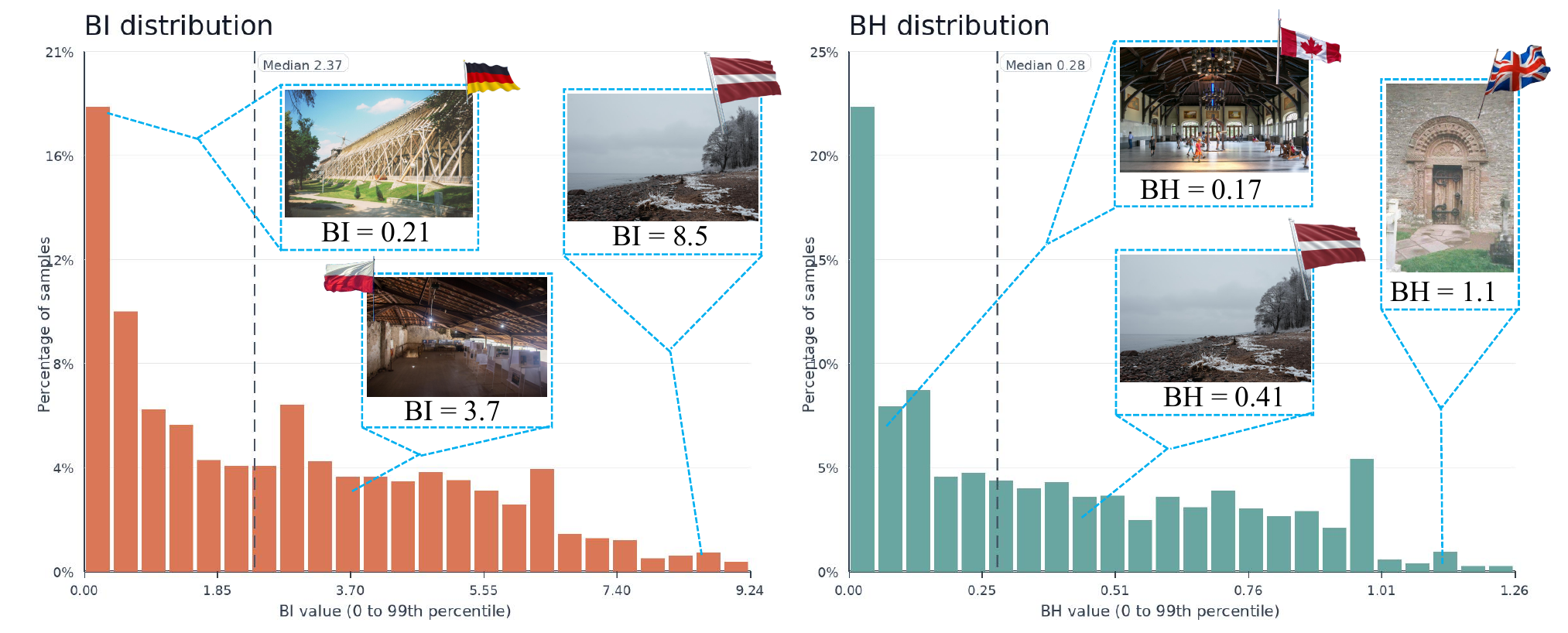}
  \vspace{-10pt}
  \caption{Empirical distributions of Bias Intensity (BI) and Bias Harmfulness (BH) in LandmarkBias-3K. The dashed lines mark the median values.}
  \label{BIBH}
\end{figure*}

In this appendix, we provide further analysis of the proposed bias metrics. 
We first examine the empirical distributions of Bias Intensity (BI) and Bias Harmfulness (BH) on LandmarkBias-3K, and then discuss a small set of counterintuitive cases with negative BI values.

Figure~\ref{BIBH} illustrates the distributions of BI and BH. 
Both exhibit clear right-skewed patterns, with most samples concentrated in low-to-moderate positive ranges and a long tail extending toward higher values. 
This suggests that while many samples exhibit moderate landmark-induced bias, a substantial portion involves stronger and more harmful effects.
The median values of BI and BH are 2.37 and 0.28, respectively, indicating that the benchmark predominantly consists of samples with positive landmark influence and non-trivial harmfulness. 
Importantly, the distribution is not dominated by extreme cases, but instead spans a wide range of bias levels. 
Overall, these results demonstrate that LandmarkBias-3K provides a diverse and challenging evaluation set, enabling robust assessment of geo-localization models under misleading landmark cues.

\section{LandmarkBias-3K Benchmark}
\label{app_sec:3k_benchmark}
In this appendix, we provide additional details on the construction of the proposed \textbf{LandmarkBias-3K} benchmark, a diagnostic dataset designed to evaluate the robustness of geo-localization models under misleading landmark cues.

\subsection{Source Data}
We construct LandmarkBias-3K from two large-scale geo-localization datasets, \textbf{MP-16}~\cite{larson2017benchmarking} and \textbf{GLDv2}~\cite{weyand2020GLDv2}, which offer broad geographic coverage and diverse real-world scenes. 
For each image, we use GroundingDINO~\cite{liu2024grounding} to detect the prominent landmark region $r$. 
Based on this region, we construct two auxiliary inputs: the landmark-removed image $x_r$, obtained by masking out $r$, and the landmark-only crop $r$, which isolates the detected landmark.

\subsection{Threshold Selection}

For each image, we perform three forward passes using the full image $x$, the landmark-removed image $x_r$, and the landmark-only region $r$. 
Following the definitions in the main text, we obtain the corresponding prediction probabilities $I_1(\cdot)$, $I_2(\cdot)$, and $I_3(\cdot)$, and compute the two bias metrics, \textbf{Bias Intensity} (BI) and \textbf{Bias Harmfulness} (BH).
These metrics capture complementary aspects of landmark-induced bias. 
A positive BI indicates that the landmark increases the model's confidence in the landmark-induced label, while a positive BH indicates that such influence is harmful, as it reduces the model's relative preference for the ground-truth label. 
We therefore retain only samples satisfying:
\begin{equation}
    BI > 0, \quad BH > 0,
\end{equation}
ensuring that the selected images exhibit both positive and harmful landmark-induced bias.
We further remove geographically ambiguous samples through manual verification. 
The resulting benchmark contains \textbf{3,000} images, forming the \textbf{LandmarkBias-3K}.

\begin{figure*}[!t]
  \centering
  \vspace{-10pt}
  \includegraphics[width=0.8\textwidth]{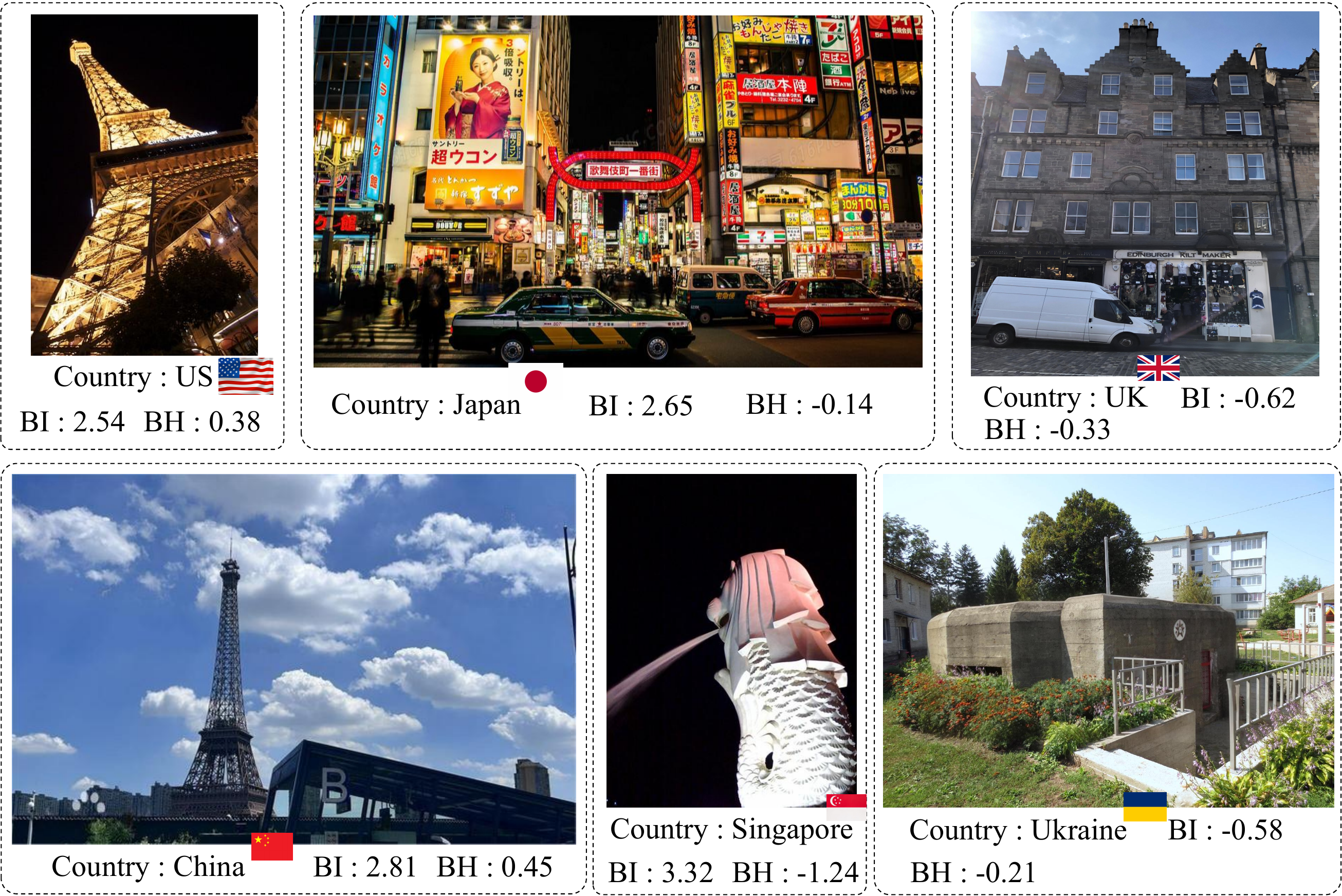}
  \vspace{-10pt}
  \caption{Representative cases with different BI/BH sign combinations: (a) $BI>0, BH>0$, harmful landmark bias; (b) $BI>0, BH<0$, strong but non-harmful landmark influence; (c) $BI<0, BH<0$, contextual reinterpretation.}
  \label{fig:more_landmark_bias_cases}
\end{figure*}

\begin{figure*}[!t]
  \centering
\includegraphics[width=0.99\linewidth]{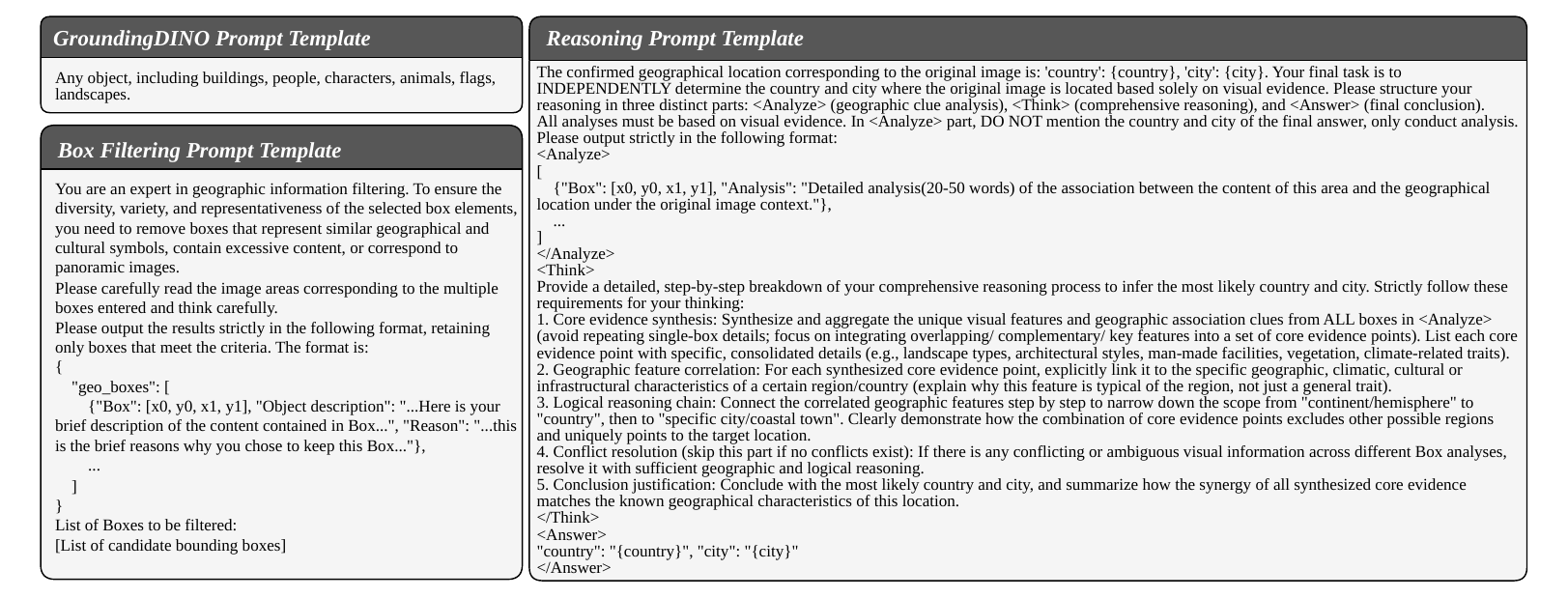}
   \vspace{-3mm}
   \caption{The prompt template used in BF-30K annotation.}
   \label{fig:Annotation_prompt}
   \vspace{-4mm}
\end{figure*}

\begin{figure*}[!t]
  \centering
   \includegraphics[width=0.95\linewidth]{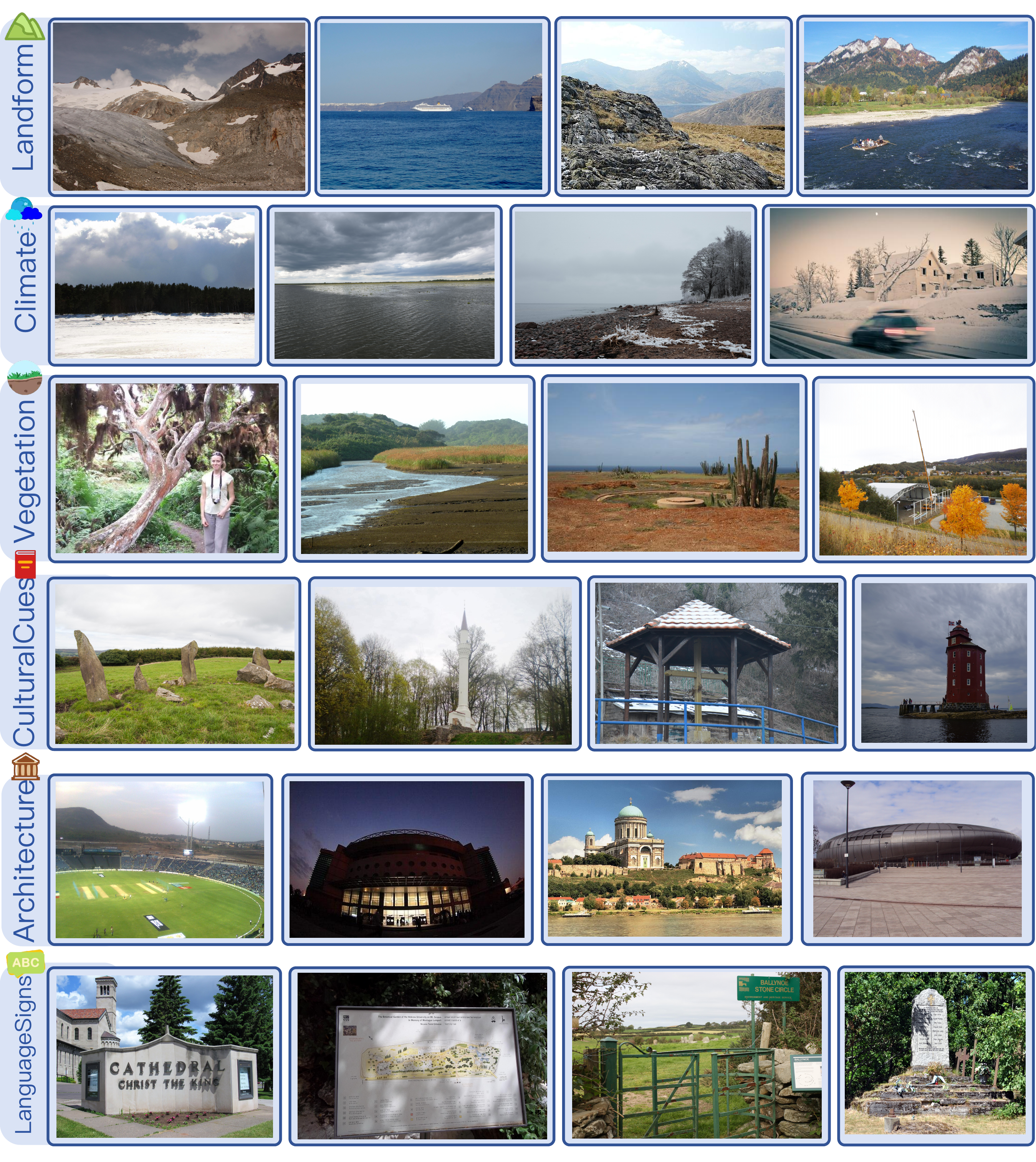}
   \vspace{-3mm}
   \caption{Representative cases of landmark categories in LandmarkBias-3K.}
   \label{fig:biasfenlei}
   \vspace{-4mm}
\end{figure*}

\subsection{More Examples}
Figure~\ref{fig:more_landmark_bias_cases} presents representative cases with different BI/BH sign combinations, illustrating that landmark cues can influence geo-localization models in qualitatively different ways.

\begin{compactitem}
    \item \textbf{(a) $\bm{BI>0}$, $\bm{BH>0}$: Harmful Landmark Bias.} 
    This case represents the primary failure mode targeted by LandmarkBias-3K. 
    In the first example, the image contains a structure resembling the Eiffel Tower, which strongly anchors the model to France when focusing on the landmark alone, while the true location is Hangzhou, China. As a result, the landmark exerts a strong positive pull toward the landmark-induced hypothesis, yielding a positive BI, and simultaneously weakens the model's relative preference for the ground truth, leading to a positive BH. 
    A similar phenomenon appears in the second example from the United States, where a highly salient landmark cue again dominates the prediction and drives the model toward an incorrect but visually plausible location hypothesis. 
    Together, these cases illustrate the standard harmful landmark-bias pattern: the landmark is visually prominent enough to attract the model, but this attraction is misleading and overrides broader contextual evidence.

    \item \textbf{(b) $\bm{BI>0}$, $\bm{BH<0}$: Strong but Beneficial Landmark Influence.} 
    In the first example, the landmark is the Merlion, a highly distinctive symbol of Singapore. The model therefore assigns strong confidence to the landmark-induced label, producing a large positive BI; however, because this landmark cue is genuinely reliable and consistent with the full scene, it supports rather than harms the correct prediction, yielding a negative BH. 
    The second example from Japan shows the same pattern: highly discriminative visual cues strongly influence the model, but this influence remains aligned with the ground-truth location. 
    These cases demonstrate that strong landmark influence is not inherently harmful. Instead, when the landmark is trustworthy and geographically discriminative, it can facilitate correct localization. They also highlight the complementary roles of BI and BH: BI measures the strength of landmark influence, while BH determines whether that influence is actually detrimental.

    \item \textbf{(c) $\bm{BI<0}$, $\bm{BH<0}$: Contextual Reinterpretation.} 
    In the first example, the model predicts Russia when observing either the landmark alone or the landmark-removed image, while the ground truth is Ukraine. However, once the full image is provided, the model shifts away from the landmark-induced prediction, resulting in a negative BI. At the same time, this shift moves the prediction relatively closer to the ground truth, yielding a negative BH. 
    A similar effect is observed in the second example from the United Kingdom, where the landmark-only interpretation is weakened after being placed back into the full scene. 
    These cases suggest that the meaning of an isolated landmark can be reinterpreted in the context of the entire image: rather than simply reinforcing the landmark-only hypothesis, the global scene can reshape or partially override it. In this sense, landmark influence is not always additive, but can depend strongly on its interaction with surrounding contextual evidence.
\end{compactitem}

Cases with $BI<0$ and $BH>0$ are relatively rare in our analysis. Overall, these examples show that landmark cues can be harmful, beneficial, or contextually corrected depending on their interaction with the surrounding visual context, which further highlights the need to jointly consider BI and BH when analyzing landmark-induced bias in geo-localization.

Figure~\ref{fig:biasfenlei} also provides representative cases from different landmark categories, such as landform, climate, vegetation, cultural cues, architecture, and language signs. This categorization shows that landmark-induced bias can originate from a diverse range of visual cues rather than only iconic landmarks, further complementing the BI/BH-based analysis above.

\begin{figure*}[!t]
  \centering
   \includegraphics[width=0.99\linewidth]{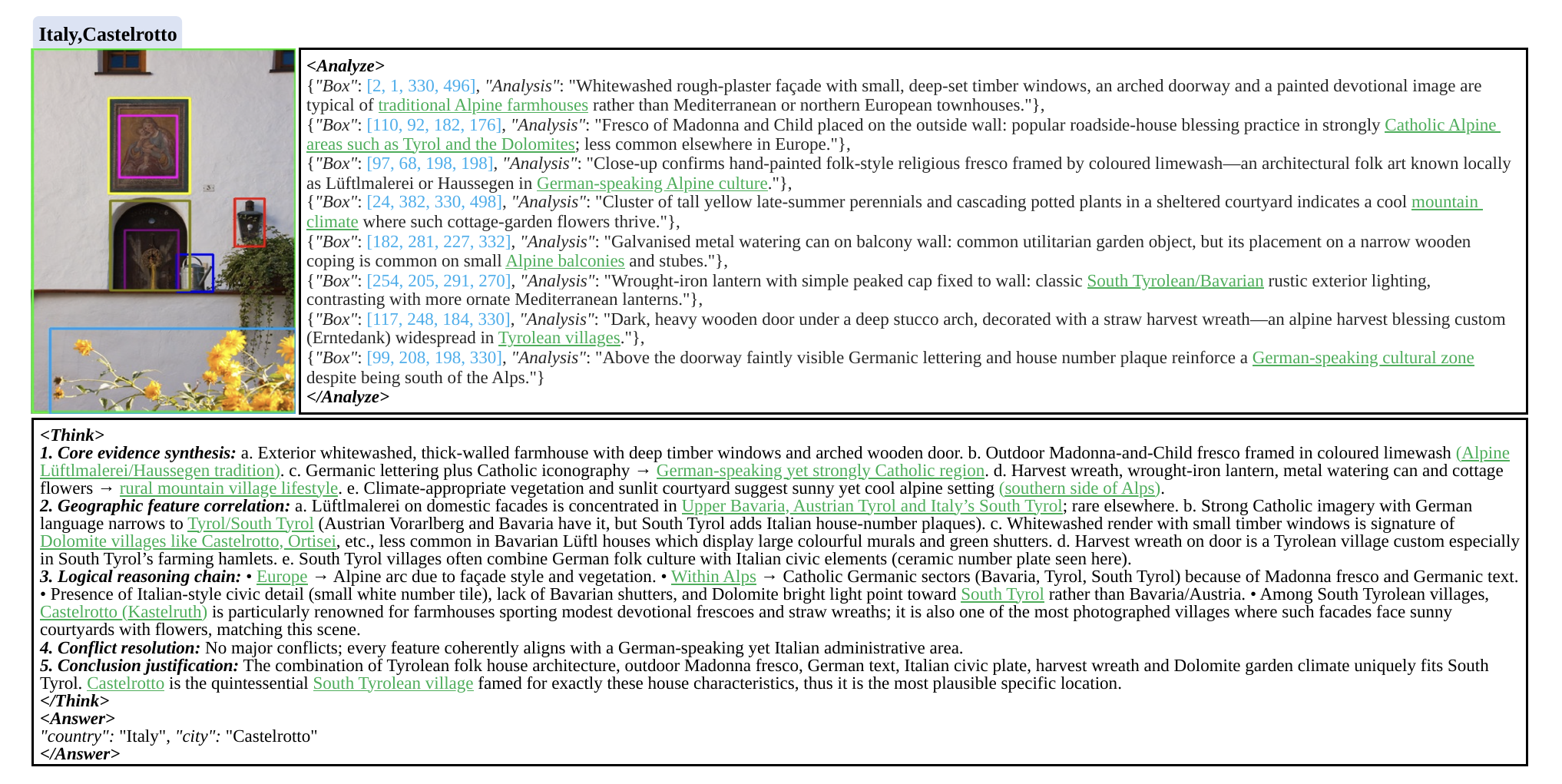}
   \vspace{-3mm}
   \caption{Visualization of an example of the training dataset.}
   \label{fig:train_example}
   \vspace{-4mm}
\end{figure*}

\begin{figure}[!t]
  \centering
   \includegraphics[width=0.99\linewidth]{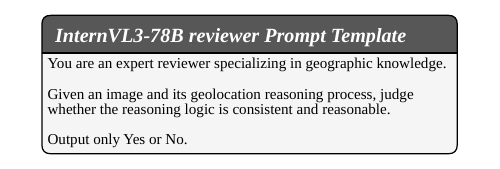}
   \vspace{-3mm}
   \caption{The prompt template used in the InternVL3-78B reviewer for multi-dimensional validation.}
   \label{fig:InternVL3_reviewer}
   \vspace{-4mm}
\end{figure}

\begin{figure*}[!t]
  \centering
   \includegraphics[width=0.99\linewidth]{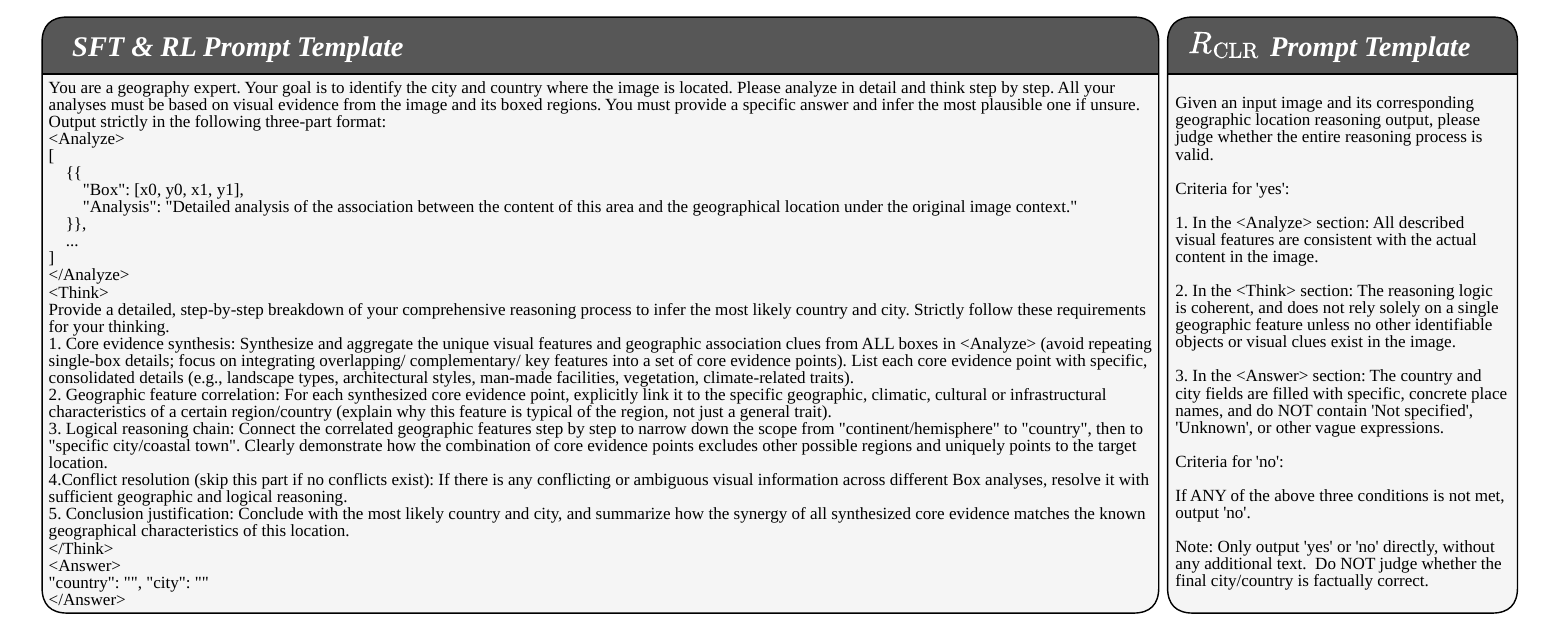}
   \vspace{-3mm}
   \caption{ Prompts in SFT and GRPO.}
   \label{fig:SFT_RL_prompt}
   \vspace{-4mm}
\end{figure*}

\section{Detailed BF-30K Annotation}
\label{app_sec:bf_30K}
Here, we detail the annotation process of the BF-30K training dataset, which consists of region proposal, semantic filtering, structured reasoning annotation, and multi-dimensional validation.

\noindent \textbf{Region Proposal.}
We employ GroundingDINO-SwinT-OGC~\cite{liu2024grounding} as an open-vocabulary object detector to identify candidate visual regions. 
To maximize the recall of geographically relevant cues, we adopt a low-threshold strategy, setting both the box confidence threshold and the text–visual alignment threshold to 0.06. 
This encourages the detection of a broad set of potential geographic elements. 
We then remove low-confidence boxes and eliminate duplicate detections using an IoU threshold of 0.75, reducing redundancy while preserving spatial coverage and semantic diversity. 
The corresponding prompt is shown in Figure~\ref{fig:Annotation_prompt}.

\noindent \textbf{Semantic Filtering.}
We further refine the candidate regions using Qwen2.5-VL-72B-Instruct~\cite{Qwen2.5-VL} as a geographic semantic filter. 
Leveraging its visual reasoning capability, the model selects regions that contain informative geographic, cultural, or environmental cues. 
The prompt used in this stage is illustrated in Figure~\ref{fig:Annotation_prompt}, where the list of candidate boxes is dynamically appended for each instance. 
This step yields a set of representative regions covering diverse geosemantic elements, including architectural structures, landmarks, vegetation, landforms, and cultural symbols.

\noindent \textbf{Structured Reasoning Annotation.}
Based on the selected regions, we perform fine-grained geographic analysis and structured reasoning annotation. 
We use Qwen2.5-VL-72B~\cite{Qwen2.5-VL} and ChatGPT-o3~\cite{openai2024o3o4mini} as reasoning models. 
The input consists of the full image and cropped patches corresponding to each region. 
Models are required to generate reasoning outputs under a strictly defined format, including regional analysis, multi-step reasoning, and final location prediction.
To ensure faithful reasoning, the prompt (Figure~\ref{fig:Annotation_prompt}) explicitly restricts the model to rely solely on visual evidence, without access to ground-truth location labels. 
This process produces structured annotations that are both interpretable and logically consistent. 
Examples of the resulting training data are shown in Figure~\ref{fig:train_example}.

\noindent \textbf{Multi-dimensional Validation.}
In this process, we employ the InternVL3-78B~\cite{zhu2025internvl3} as an independent verification model to conduct cross-model consistency checks on the reasoning annotations.
The prompt used for the model is illustrated in Figure~\ref{fig:InternVL3_reviewer}, whose core task is to judge whether the geolocation reasoning process is logically self-consistent and aligned with visual content based on the image, and output only "Yes" or "No" as the decision result.
After automated validation, samples identified as contradictory or inconsistent are filtered out.
In addition, we perform manual spot checks on the entire annotated dataset with a sampling ratio of 5\% to further correct annotation errors, place name errors, logical conflicts, and hallucinations that may be missed by model-based verification.

\section{Training Prompt Template}
To ensure input consistency between fine-tuning and testing, we design and employ a unified prompt template for vision-language model, as illustrated on the left side of Figure~\ref{fig:SFT_RL_prompt}, which guarantees a fair comparison of model performance. Meanwhile, the prompt template for the Comprehensive Logical Reasoning Reward ($R_{\text{CLR}}$) is shown on the right side of Figure~\ref{fig:SFT_RL_prompt}.

\section{Detailed Experimental Settings}
\label{app_sec:settings}

\subsection{Detailed Evaluation Benchmarks}
\label{app_sec:detailed_benchmarks}

We evaluate our method on four geo-localization benchmarks with different data sources and visual characteristics.

\textbf{IM2GPS}~\cite{hays2008im2gps} is a classical geo-localization benchmark. 
Its standard test set contains 237 images drawn from a large Flickr-based geo-tagged image collection. 
Despite its relatively small size, many samples are visually distinctive and suitable for geographic reasoning, making it a widely used baseline for early evaluation.

\textbf{IM2GPS3K} is a larger benchmark introduced by~\citet{vo2017revisiting} to enable more robust evaluation. 
It contains 3,000 images sampled from the original IM2GPS collection, providing a broader and more challenging test set. 
Compared to the standard IM2GPS split, it exhibits greater diversity in scene types, including landmarks, urban environments, and natural landscapes.

\textbf{YFCC4K}~\cite{vo2017revisiting} consists of 4,000 randomly sampled images from the YFCC-100M dataset~\cite{thomee2016yfcc100m}, a large-scale Flickr-based multimedia collection. 
As a general-purpose dataset, YFCC4K is visually diverse and less biased toward iconic landmarks, making it a complementary benchmark for evaluating generalization beyond landmark-centric scenarios.

\textbf{Landmark-Bias-3K Benchmark} is our proposed diagnostic benchmark for evaluating landmark-induced bias. 
It contains 3,000 images explicitly curated to assessing whether models over-rely on dominant landmark evidence.

\begin{table}[!h]
  \centering
  \caption{The values of each hyperparameter.}
  \label{tab:hyperparameters}
  \vspace{-2mm}
  \begin{tabular}{llc}
    \toprule
    \textbf{Symbol} & \textbf{Description} & \textbf{Value} \\
    \midrule
    \multicolumn{3}{c}{\textit{Supervised Fine-Tuning (SFT) Configuration}} \\
    \midrule
    $l_r$ & Initial learning rate & $1\times10^{-4}$ \\
    $r_{\text{lora}}$ & LoRA rank & 64 \\
    $\alpha_{\text{lora}}$ & LoRA alpha & 128 \\
    $p_{\text{lora}}$ & LoRA dropout & 0.05 \\
    $b_{\text{train}}$ & Per-device training batch size & 1 \\
    $L_{\text{max}}$ & Maximum sequence length & 8192 \\
    $r_{\text{warm}}$ & Warmup ratio & 0.05 \\
    \midrule
    \multicolumn{3}{c}{\textit{GRPO Reinforcement Learning Configuration}} \\
    \midrule
    $l_{r,\text{rl}}$ & Initial learning rate & $1\times10^{-5}$ \\
    $G$ & Group size & 8 \\
    $t$ & Sampling temperature & 0.7 \\
    $\beta$ & KL divergence coefficient & 0.001 \\
    $r_{\text{lora,rl}}$ & LoRA rank & 64 \\
    $\alpha_{\text{lora,rl}}$ & LoRA alpha & 128 \\
    $p_{\text{lora,rl}}$ & LoRA dropout & 0.05 \\
    $L_{\text{max,rl}}$ & Maximum sequence length & 8192 \\
    $L_{\text{comp}}$ & Maximum completion length & 1024 \\
    $r_{\text{warm,rl}}$ & Warmup ratio & 0.01 \\
    $\lambda_1$ & Weight of $R_{\text{geo}}$ & 1.2 \\
    $\lambda_2$ & Weight of $R_{\text{box}}$ & 0.3 \\
    $\lambda_3$ & Weight of $R_{\text{CLR}}$ & 0.5 \\
    $\lambda_{\text{fmt}}$ & Weight of $R_{\text{format}}$ & 1.0 \\
    $\alpha$ & City-level weight in $R_{\text{geo}}$ & 0.8 \\
    $\tau_{\text{IoU}}$ & IoU threshold in $R_{\text{box}}$ & 0.6 \\
    \midrule
    \multicolumn{3}{c}{\textit{LLM Judge Configuration}} \\
    \midrule
    $t_{\text{judge}}$ & Sampling temperature & 0.2 \\
    $L_{\text{max,judge}}$ & Maximum generation tokens & 5 \\
    $l_{\text{prob}}$ & Return log probabilities & True \\
    $k_{\text{top}}$ & Number of top logprobs & 5 \\
    \bottomrule
  \end{tabular}%
  \vspace{-4mm}
\end{table}

\begin{figure*}[!t]
  \centering
   \includegraphics[width=0.99\linewidth]{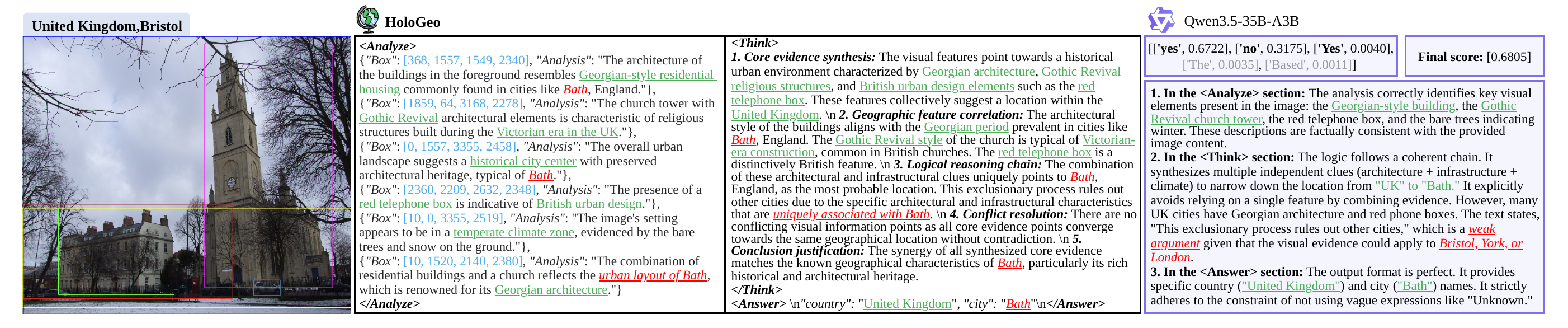}
   \vspace{-2mm}
   \caption{A case of $\mathcal{R}_{\text{CLR}}$.}
   \label{fig:CLR_case}
\end{figure*}

\subsection{Detailed Implementation}

We conduct SFT as the first stage of model training, with GRPO-based post-training. 
We adopt Qwen2.5-VL-7B-Instruct as the base model and perform parameter-efficient fine-tuning using LoRA with a rank of 64. During SFT, we train the model for 1 epoch using the AdamW optimizer with an initial learning rate of $1\times10^{-4}$, a warmup ratio of 0.05. Training is deployed on 4$\times$ NVIDIA H100 GPUs with DeepSpeed Zero-3 for memory optimization, yielding a global batch size of 32. For GRPO-based reinforcement learning, we use a smaller learning rate of $1\times10^{-5}$, a warmup ratio of 0.01, and set weight decay to 0.1. We keep the LoRA configuration consistent across both training stages. We employ a cosine learning rate scheduler in both training phases. 
We summarize the key hyperparameters used for SFT and GRPO in our training setup in Table \ref{tab:hyperparameters}. 
These settings are chosen based on standard practices for fine-tuning large vision-language models and further adjusted via preliminary ablation studies conducted on a held-out validation set. 
Unless stated otherwise, the same configuration is adopted across all experiments to ensure comparability and reproducibility.

Additionally, Figure~\ref{fig:CLR_case} presents an example of the proposed $\mathcal{R}_{\text{CLR}}$, illustrating how Qwen3.5-35B-A3B evaluates a generated geographic reasoning chain.
The reward model performs structured assessment across the three reasoning stages (\texttt{<Analyze>}, \texttt{<Think>}, and \texttt{<Answer>}), jointly considering visual grounding accuracy, logical consistency, and answer specificity. 
By explicitly verifying whether the reasoning is supported by localized visual evidence and whether multiple cues are coherently integrated, the model can effectively identify flawed or over-simplified reasoning patterns.
Moreover, the probabilistic scoring mechanism provides a soft and stable measure of reasoning validity, enabling fine-grained optimization during reinforcement learning. 
This design allows $\mathcal{R}_{\text{CLR}}$ to serve as a reliable signal for guiding the model toward factually grounded and logically consistent multi-evidence reasoning.

\section{Extended Experiments}
\label{app_sec:experimental_results}

\subsection{Analysis of Landmark Bias}
\begin{figure*}[!t]
  \centering
   \includegraphics[width=0.95\linewidth]{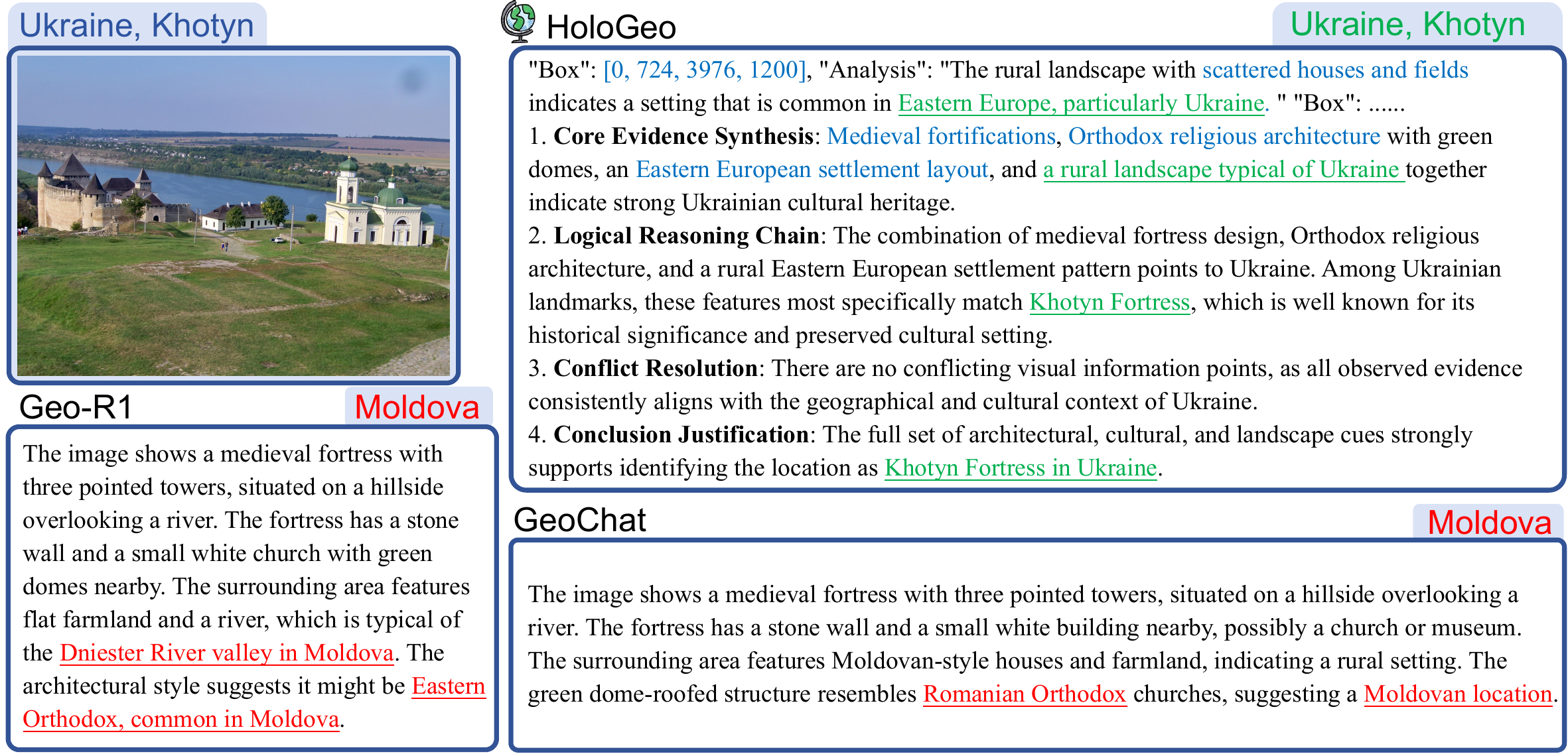}
   \vspace{-2mm}
   \caption{Case study: Reasoning comparison of three different models on the same input image. Reliable geographic information identified by the models is annotated in the text.}
   \label{fig:case_study2}
\end{figure*}

Figure~\ref{fig:case_study2} presents another representative example of \emph{landmark-induced bias} in image geo-localization. 
The scene contains several visually salient yet geographically ambiguous cues, including a medieval fortress, green-domed Orthodox architecture, and a riverfront landscape, which collectively evoke a broad Eastern European visual prior. 
As shown, both Geo-R1 and GeoChat are attracted by these dominant but insufficient landmark cues and prematurely commit to \textit{Moldova}, producing plausible yet incorrect hypotheses based on coarse landmark-level similarity rather than robust geographic evidence. 
In contrast, our model remains less susceptible to such dominant visual bias: instead of over-relying on a single prominent structure or stylistic cue, it integrates more holistic scene context, aggregating complementary evidence from settlement layout, architectural composition, and surrounding landscape, which leads to the correct localization in \textit{Khotyn, Ukraine}.

\subsection{BI and BH Distribution Analysis}

\begin{figure}[!t]
  \centering
   \includegraphics[width=0.92\linewidth]{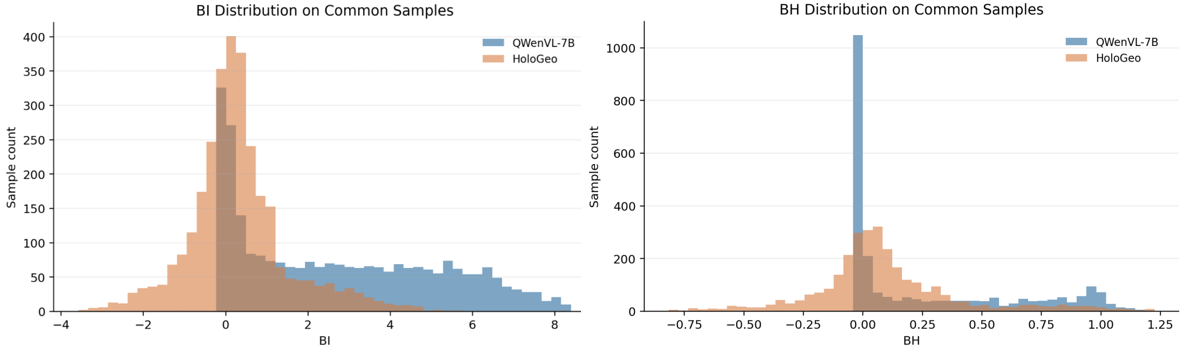}
   \vspace{-2mm}
   \caption{Comparison of the BI and BH distributions between
  Qwen2.5-VL-7B-Instruct and HoloGeo on LandmarkBias-3K.}
   \label{fig:bibh_compare}
\end{figure}
To further assess whether HoloGeo mitigates landmark bias at the model-behavior level, we compute BI and BH for Qwen2.5-VL-7B-Instruct and HoloGeo on LandmarkBias-3K. As shown in Figure~\ref{fig:bibh_compare}, compared with Qwen2.5-VL-7B-Instruct, HoloGeo exhibits BI and BH distributions that are more concentrated in lower-value regions, with markedly shorter high-value tails. This indicates that HoloGeo reduces excessive landmark anchoring and mitigates the harmful influence of misleading landmarks on accurate localization, further validating its effectiveness in alleviating landmark bias.

\subsection{More Case Examples}
\label{app_sec:more_case_examples}

\begin{figure*}[!t]
  \centering
   \includegraphics[width=0.92\linewidth]{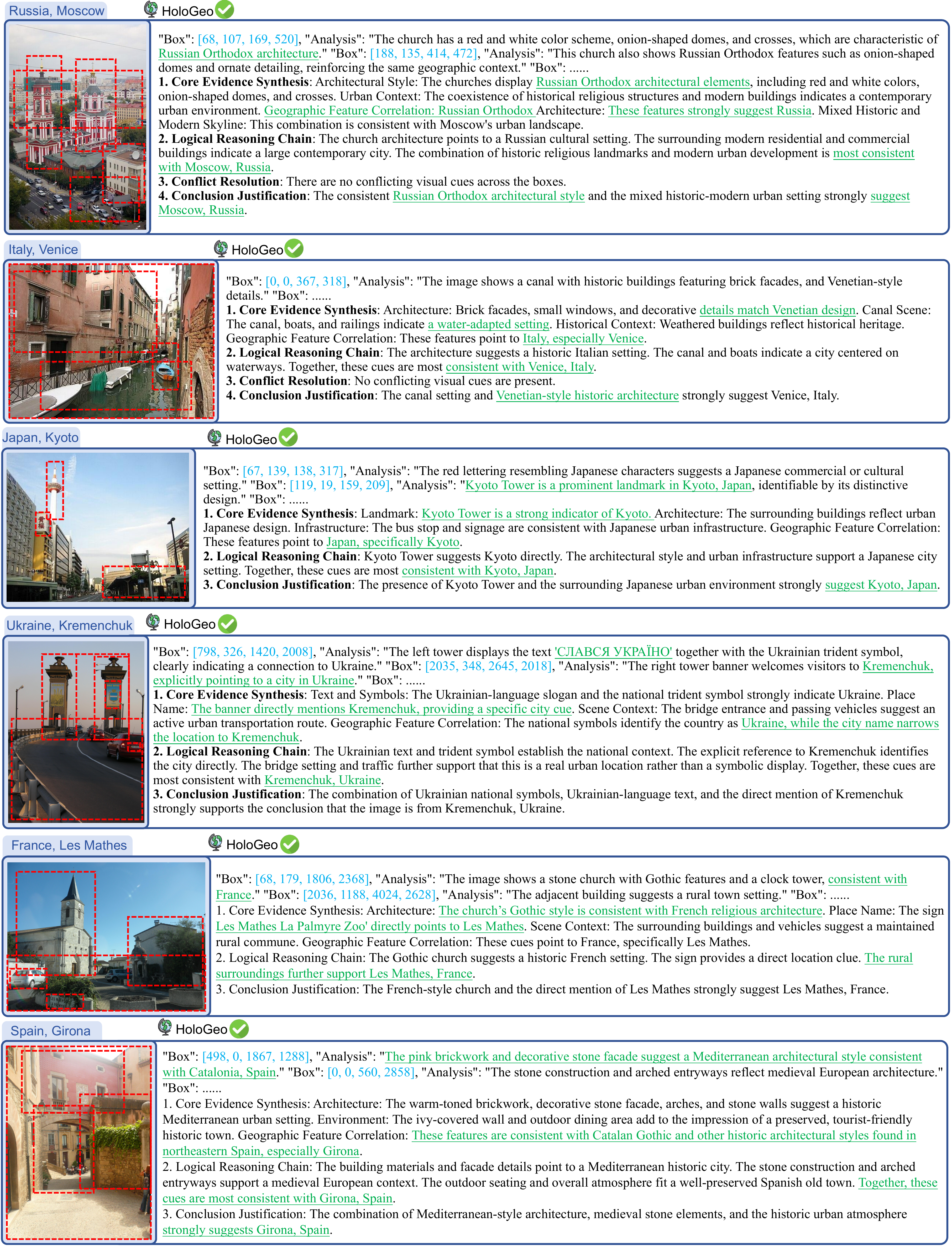}
   \vspace{-2mm}
   \caption{More correct cases of HoloGeo on LandmarkBias-3K.}
   \label{fig:landmarkbias_1}
\end{figure*}

\begin{figure*}[!t]
  \centering
   \includegraphics[width=0.92\linewidth]{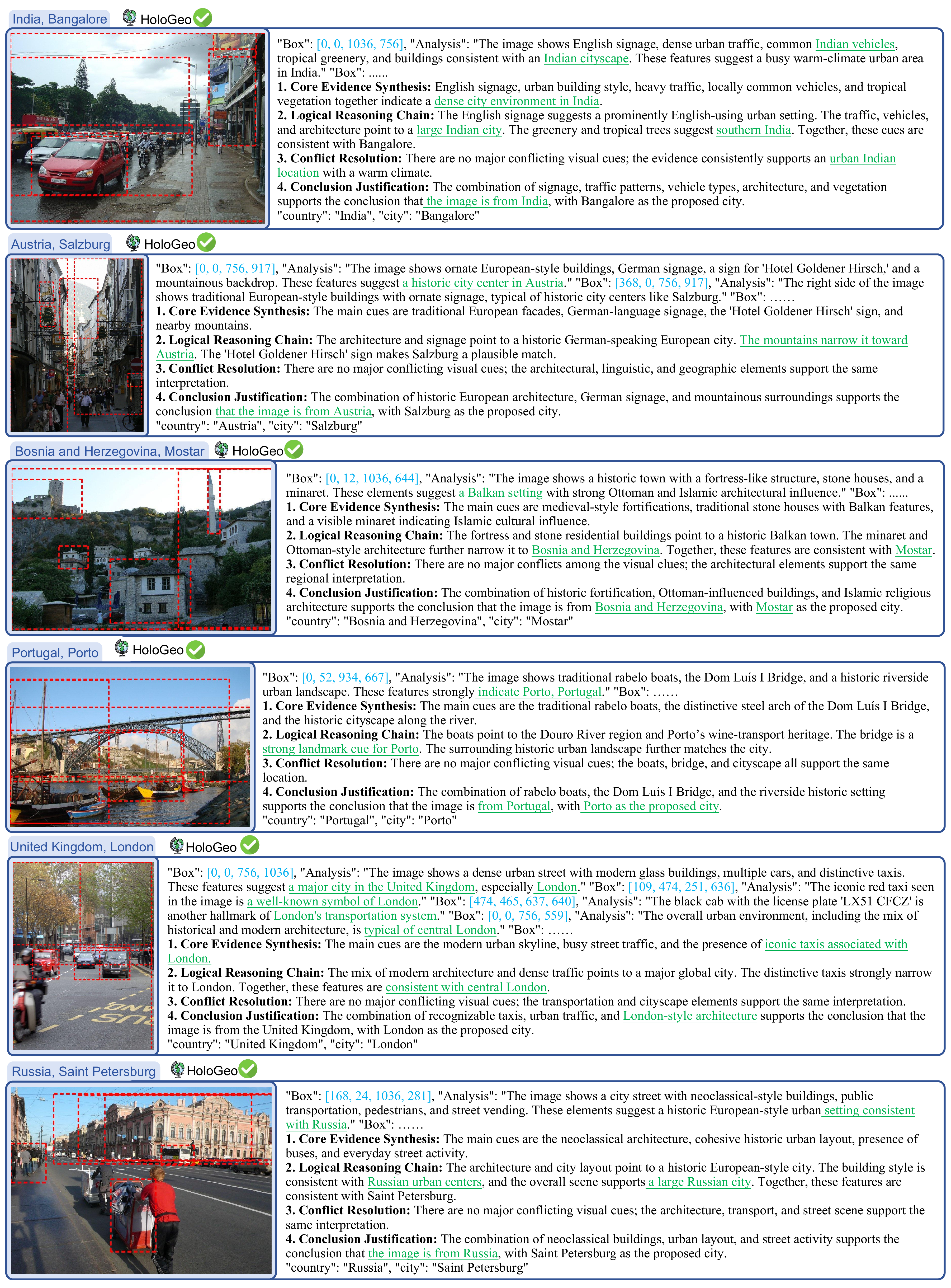}
   \vspace{-2mm}
   \caption{More correct cases of HoloGeo on IM2GPS.}
   \label{fig:landmarkbias_im2gps}
\end{figure*}

\begin{figure*}[!t]
  \centering
   \includegraphics[width=0.92\linewidth]{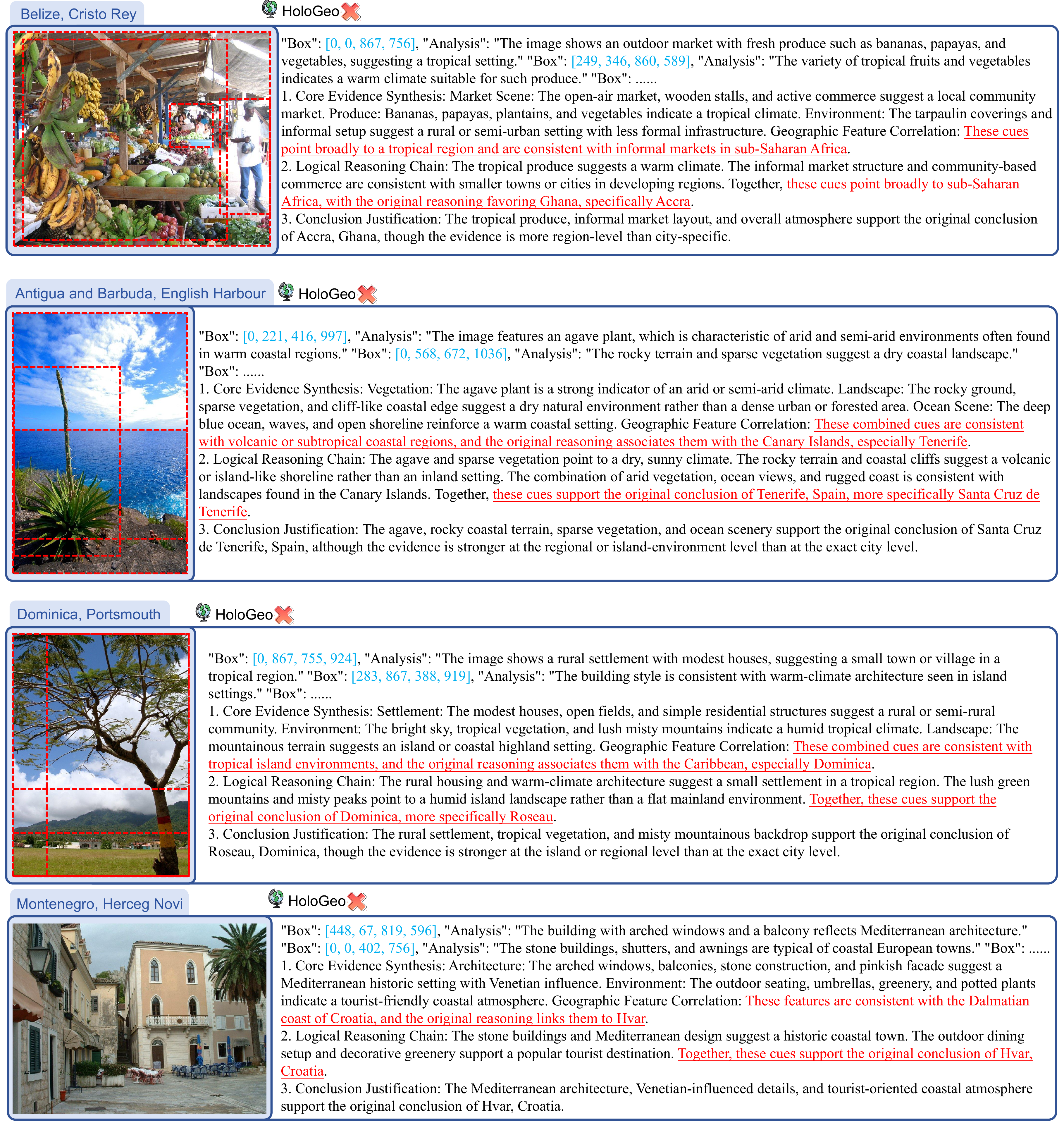}
   \vspace{-2mm}
   \caption{Some failure cases of HoloGeo on IM2GPS.}
   \label{fig:im2gps_error}
\end{figure*}

Figures~\ref{fig:landmarkbias_1} and~\ref{fig:landmarkbias_im2gps} present additional successful examples of HoloGeo on LandmarkBias-3K and IM2GPS, respectively. 
Across these cases, the model does not rely solely on a single salient landmark or iconic structure.
Instead, it integrates broader scene evidence, such as architectural style, urban or settlement layout, surrounding vegetation, traffic patterns, textual signs, and environmental context, and then combines these complementary cues to infer the correct location. 
This further supports our main claim that HoloGeo is able to mitigate over-reliance on dominant landmark cues and make more robust geo-localization decisions through evidence-driven reasoning.
Figure~\ref{fig:im2gps_error} shows several failure cases on IM2GPS. A common pattern is that these examples come from relatively uncommon or underrepresented countries or regions, where the visual evidence is less frequently observed during training. In such cases, although the model can still capture some coarse regional characteristics, its geographic knowledge is insufficient to reliably distinguish rare locations from more familiar visually similar ones, which leads to incorrect predictions. These results indicate that, beyond reducing landmark bias, further improvement also requires broader and more fine-grained geographic knowledge coverage for long-tail locations.

\begin{figure*}[!t]
  \centering
   \includegraphics[width=0.8\linewidth]{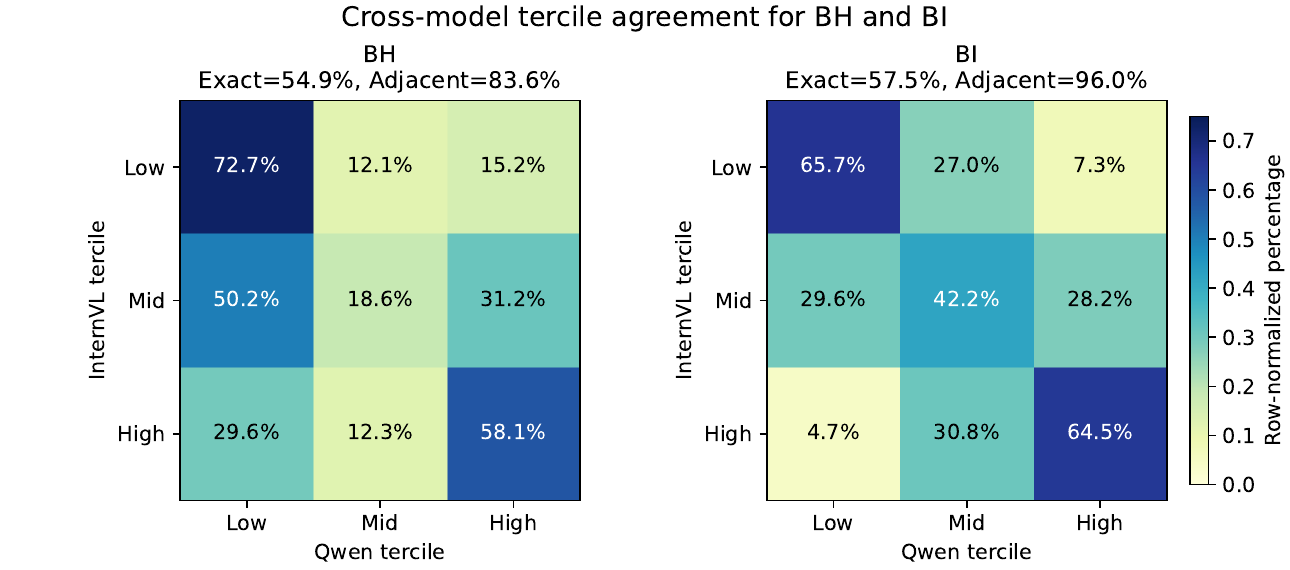}
   \vspace{-2mm}
   \caption{Cross-model validation of BH and BI based on tercile agreement between InternVL3-78B and Qwen2.5-VL-72B. The confusion matrices show that coarse relative trends are largely consistent across model families.}
   \label{fig:model_cross_validation}
\end{figure*}

\section{Model Cross-Validation}
\label{app_sec:jcyz}
To examine whether the proposed BH and BI signals are tied to a specific backbone, we compare the predictions of InternVL3-78B~\cite{zhu2025internvl3} and Qwen2.5-VL-72B~\cite{Qwen2.5-VL} on the full evaluation set (Figure~\ref{fig:model_cross_validation}). 
Since different VLMs may produce scores on different numeric scales, we assess cross-model consistency at the level of relative trends rather than raw score values. 
Specifically, for each metric (BH or BI), we partition each model’s predictions into three equal-frequency bins based on its own score distribution, corresponding to low, medium, and high terciles.

Based on these tercile assignments, we compute two agreement measures. 
The \emph{exact-match ratio} is the proportion of samples assigned to the same tercile by both models. 
The \emph{adjacent-or-better ratio} counts samples whose assignments differ by at most one tercile, treating only low-versus-high cases as strong disagreements. 
This formulation captures whether the two models preserve a consistent coarse ordering of samples despite differences in score scale.
As shown in Figure~\ref{fig:model_cross_validation}, both metrics exhibit substantial cross-model consistency. 
For BH, 54.88\% of samples fall into the same tercile and 83.62\% into the same or an adjacent tercile. 
For BI, the agreement is even stronger, with 57.46\% exact matches and 96.00\% adjacent-or-better consistency. 
These values are significantly higher than a random assignment baseline, indicating that the relative ordering induced by BH and BI is largely preserved across model families.
The row-normalized confusion matrices further illustrate this behavior. 
Each row corresponds to a tercile assigned by InternVL and each column to a tercile assigned by Qwen. 
Diagonal entries indicate exact agreement, near-diagonal entries reflect mild shifts, and corner entries correspond to cross-extreme disagreements. 
The matrices show that most discrepancies occur between neighboring bins, while extreme contradictions (low vs. high) are rare, particularly for BI.

Overall, these observations suggest that the BH and BI patterns capture stable, model-agnostic patterns of landmark-induced bias, rather than artifacts specific to a particular VLM backbone.

\section{Limitation}
Although our proposed HoloGeo model has achieved excellent geolocalization performance on multiple public datasets, it still has certain limitations.
First, since the training data mainly comes from existing datasets such as MP-16~\cite{larson2017benchmarking} and GLDv2~\cite{weyand2020GLDv2}, the data distribution is somewhat biased, resulting in poor recognition performance of the model in remote areas, underrepresented landforms, and data-sparse regions.
Second, the current model still relies on supervised training with annotated datasets, which incurs certain training costs.
In addition, the social impact of geolocalization models deserves equal attention.
Caution should be exercised in real-world applications to guard against potential privacy leakage risks.
The usage must be strictly regulated to prevent it from being used for illegal tracking, facilitating criminal activities, or violating personal privacy and spatial security.

\end{document}